%% file: main.tex
\documentclass[10pt,twocolumn,letterpaper]{article}

\usepackage[accsupp]{axessibility}
\usepackage{cvpr}              
\input{preamble}
\definecolor{cvprblue}{rgb}{0.21,0.49,0.74}
\usepackage[pagebackref,breaklinks,colorlinks,allcolors=cvprblue]{hyperref}

\def\paperID{45622} 
\def\confName{CVPR}
\def\confYear{2026}

\usepackage{multirow}

\title{AVGGT: Rethinking Global Attention for Accelerating VGGT}

\author{
    Xianbing Sun$^{1,2}$\footnotemark[1] \quad
    Zhikai Zhu$^{1}$\footnotemark[1] \quad
    Zhengyu Lou$^{1}$\footnotemark[1] \quad
    Bo Yang$^{2}$\\
    Jinyang Tang$^{2}$ \quad
    Liqing Zhang$^{1}$\footnotemark[2] \quad
    He Wang$^{2}$\footnotemark[2] \quad
    Jianfu Zhang$^{1}$\footnotemark[2]\\[2mm]
    $^{1}$Shanghai Jiao Tong University,
    $^{2}$Ant Group \\
    \texttt{\{fufengsjtu, lqzhang, c.sis\}@sjtu.edu.cn, he.wang@antgroup.com} \\
}

\begin{document}

\twocolumn[{
\renewcommand\twocolumn[1][]{#1}%
\maketitle
\begin{center}
  \includegraphics[width=\textwidth]{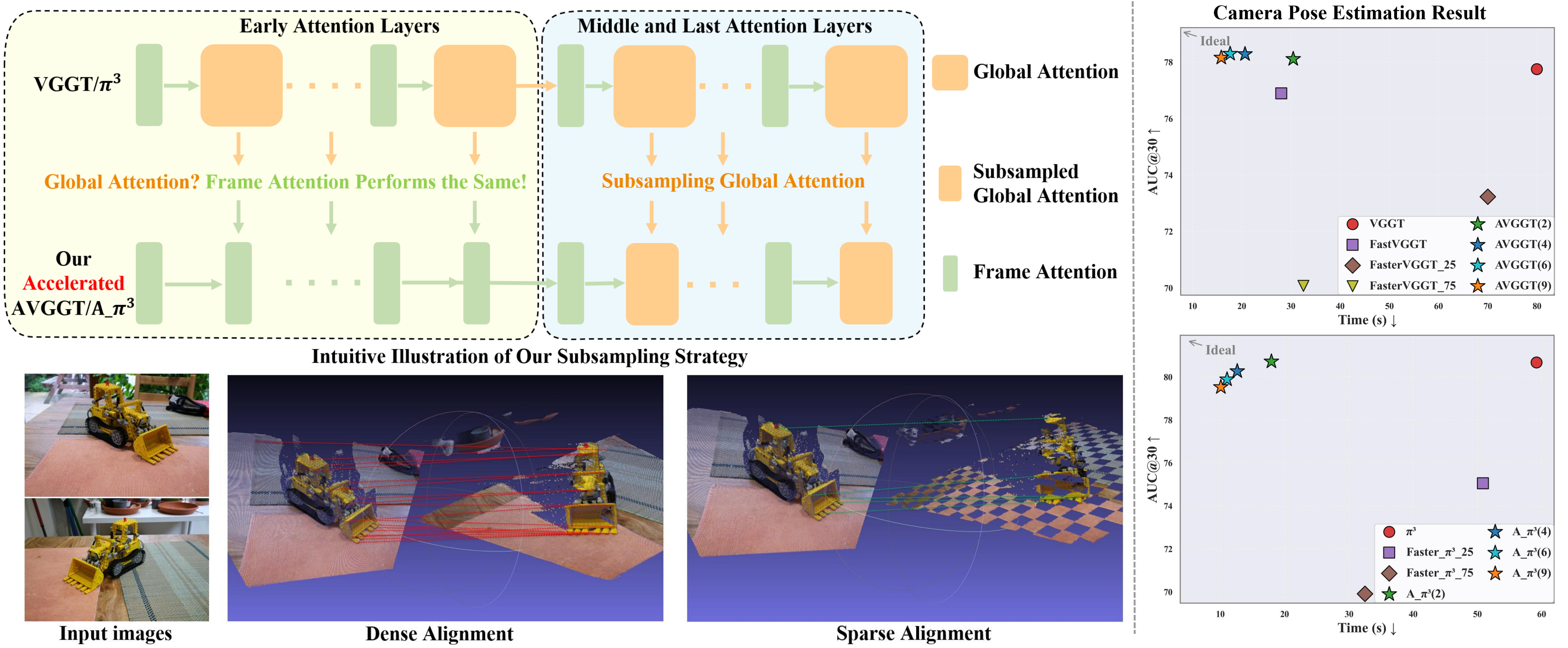}
  \captionof{figure}{
    \textbf{Left:} Overview of our acceleration pipeline and the intuition behind subsampling global attention. Our analysis shows that global attention in VGGT mainly performs alignment by identifying spatially corresponding regions across views, which can be preserved using only a sparse set of Key/Value tokens while keeping all Queries. This motivates our uniform grid-based K/V subsampling strategy. \textbf{Right:} Camera pose estimation results on the 7-Scenes dataset under the dense setting (each test sequence contains 333 frames). AVGGT and A\_\(\pi^3\)  denote our methods, where the number in parentheses (\textit{e.g.}, 2, 4, 6, 9) indicates the subsampling factor applied to the global attention layers.
    }
  \label{fig:teaser}
\end{center}
}]

\begingroup
\renewcommand{\thefootnote}{\fnsymbol{footnote}} 
\footnotetext[1]{Equal contribution.}
\footnotetext[2]{Corresponding authors.}
\endgroup

\input{section/0_abstract}    
\input{section/1_intro}
\input{section/2_rela_work}
\input{section/3_analysis}
\input{section/4_method}
\input{section/5_experiments}

\input{section/6_conclusion}

{
    \small
    \bibliographystyle{ieeenat_fullname}
    \bibliography{main}
}
\input{section/7_supp}

\end{document}

%% file: section/0_abstract.tex
\begin{abstract}
Models such as VGGT and $\pi^3$ have shown strong multi-view 3D performance, but their heavy reliance on global self-attention results in high computational cost.
Existing sparse-attention variants offer partial speedups, yet lack a systematic analysis of how global attention contributes to multi-view reasoning.
In this paper, we first conduct an in-depth investigation of the global attention modules in VGGT and $\pi^3$ to better understand their roles. 
Our analysis reveals a clear division of roles in the alternating global-frame architecture: early global layers do not form meaningful correspondences, middle layers perform cross-view alignment, and last layers provide only minor refinements.
Guided by these findings, we propose a training-free two-step acceleration scheme: (1) converting early global layers into frame attention, and (2) subsampling global attention by subsampling K/V over patch tokens with diagonal preservation and a mean-fill component.
We instantiate this strategy on VGGT and $\pi^3$ and evaluate across standard pose and point-map benchmarks. Our method achieves substantial inference acceleration across different context lengths, yielding about $2\times$ speedup at 100 frames, $4$--$5\times$ at 300 frames, and $8$--$10\times$ at 800 frames, while matching or slightly improving the accuracy of the original models and remaining robust in extremely dense multi-view settings where prior sparse-attention baselines fail.
\end{abstract}

%% file: section/1_intro.tex
\section{Introduction}
\label{sec:intro}

In the field of 3D vision, many fundamental tasks have been extensively studied over the years, including 3D reconstruction, depth estimation, pose estimation, and point tracking. 
These tasks are crucial for various real-world applications, such as autonomous driving and AR/VR.  
Classical pipelines like COLMAP~\cite{schonberger2016structure,schonberger2016pixelwise} detect and match keypoints~\cite{lindenberger2021pixel} with descriptors such as SIFT~\cite{lowe2004distinctive} to build explicit cross-view correspondences, but often fail in weakly textured regions~\cite{tombari2013bold}.
DUSt3R~\cite{wang2024dust3r}, built on CroCo~\cite{weinzaepfel2022croco}, learns cross-view correlations via cross-attention and directly predicts dense 3D point maps from two images with strong generalization.
However, it only outputs point maps, so pose estimation still falls back to classical pipelines, and extending to multi-view settings requires expensive global alignment.
VGGT~\cite{wang2025vggt} integrates multiple tasks in a single transformer with heads for depth, point maps, pose, and tracking, and accepts many input images. An ablation over attention schemes shows that global self-attention outperforms cross-attention, and that alternating global self-attention with frame self-attention yields the best results.
This design has influenced subsequent models, including $\pi^3$~\cite{wang2025pi} and MapAnything~\cite{keetha2025mapanything}, which also adopt global self-attention to model multi-view correlations.

Although VGGT demonstrates that alternating attention works well, it raises two key questions: 
\textit{\textbf{(Q1)} What is the underlying mechanism behind alternating attention that makes it effective?}  
\textit{\textbf{(Q2)} Given that global self-attention is computationally expensive, can we reduce its cost without sacrificing performance?}
For \textbf{(Q1)}, we perform a layer-wise analysis of global attention in VGGT and $\pi^3$.
We find a clear division of roles in the alternating architecture: early global layers do not form meaningful correspondences since features at this stage lack sufficient 3D information, middle layers perform cross-view alignment by linking spatially corresponding tokens, and last layers provide only minor refinements.  
Together with the interleaved frame-attention blocks, this yields an iterative mechanism in which global attention enforces multi-view consistency while frame attention refines per-view structure.
For \textbf{(Q2)}, since global self-attention in VGGT introduces a significant computational cost of \(O(N^2)\) for \(N\) frames, several methods such as FastVGGT~\cite{shen2025fastvggt} and FasterVGGT~\cite{wang2025faster} have attempted to mitigate this cost by adopting sparse-attention mechanisms from other domains, including token merging~\cite{bolya2022token,haurum2023tokens,renggli2022learning,zeng2022not} and block-sparse attention~\cite{dao2021pixelated}.  
However, these approaches typically lack a systematic analysis of the complete forward process in VGGT and do not exploit the alignment-centric nature of global attention.
Motivated by this gap, and guided by our findings for \textbf{Q1}, we propose a training-free two-step acceleration scheme for VGGT-style models:
(i) Since the early global attention layers contribute little to multi-view consistency, we convert them to frame attention.
(ii) As the remaining global layers mainly build multi-view correspondences by aligning spatially corresponding tokens, we view them through a point-cloud lens: aligning two point clouds with a rigid transform in principle requires only a few anchor points, making dense matching redundant.  
This insight leads to an aggressive but simple subsampling of global attention, where we uniformly subsample patch tokens as Keys/Values on a 2D grid (\textit{e.g.}, one token per $s_h\times s_w$ window) while keeping all Queries and special tokens.

In this paper, we implement our strategy on VGGT and its variant \(\pi^3\), both of which achieve substantial inference acceleration with almost no performance degradation.  
For instance, our method yields up to \(8\times\) and \(10\times\) speedups on VGGT and \(\pi^3\), respectively, when processing 800 input images.  
These results not only improve practical efficiency but also empirically validate our analysis and hypotheses about the role of global attention.
Overall, our main contributions are:
(i) We conduct a detailed layer-wise analysis of global attention in VGGT and $\pi^3$, revealing the specific roles of early, middle, and last global layers and explaining why alternating global and frame attention is effective.  
(ii) We propose a training-free two-step acceleration pipeline derived from a 3D alignment perspective: Global-to-Frame conversion for early layers and a subsampling strategy for global attention with diagonal preservation.
(iii) Extensive experiments show that our method achieves up to \textit{$8$-$10\times$ speedup} while \textit{matching or slightly improving the original models} across both sparse and dense multi-view settings.

%% file: section/2_rela_work.tex
\section{Related Work}
\label{sec:rela}

\noindent\textbf{Feed-Forward 3D Reconstruction.}
Traditional 3D reconstruction methods such as Structure-from-Motion (SfM)~\cite{cui2017hsfm,hartley2003multiple,pan2024global,schonberger2016structure} typically consist of multiple stages, including feature extraction and matching, triangulation, and bundle adjustment. 
Multi-View Stereo (MVS)~\cite{furukawa2015multi,schonberger2016pixelwise} is usually built upon the results of SfM to reconstruct dense geometry. 
The key in this classical pipeline is to extract and match keypoints across different views. 
A representative method is COLMAP, which is widely used in industry. 
However, such approaches suffer from several issues, including failures in weakly textured regions and the accumulation of errors across multiple processing stages.  

A major breakthrough toward learning-based 3D reconstruction is DUSt3R~\cite{wang2024dust3r}, which is built upon CroCo~\cite{weinzaepfel2022croco}. 
CroCo is trained to establish cross-view correlations through cross-attention. 
Intuitively, this serves as the learned counterpart of keypoint matching in traditional 3D reconstruction, which explains why DUSt3R can directly infer dense 3D point maps from two images while maintaining spatial consistency.

VGGT~\cite{wang2025vggt} further extends this idea by scaling up the model and supporting multiple input images. 
It jointly predicts camera poses, depth maps, point maps, and tracking results within a unified transformer-based framework. 
More importantly, it reveals that using global self-attention to model multi-view correlations yields better performance. 
Its successor, \(\pi^3\)~\cite{wang2025pi}, goes one step further: unlike VGGT, which predicts all point maps in the coordinate system of a reference view, \(\pi^3\) removes the reference view constraint and achieves permutation invariance by eliminating camera embeddings and modifying the training loss.  
Notably, both VGGT and \(\pi^3\) share a similar general architecture: they encode input images using a pretrained DINOv2 encoder and employ alternating global and frame self-attention layers to build global feature representations. 
While effective, the computational cost introduced by global self-attention becomes substantial as the number of input images increases.

\noindent\textbf{Accelerating VGGT.}
To alleviate the computational cost introduced by global attention, VGGT Long~\cite{deng2025vggt} does not process the entire input image sequence at once. 
Instead, it partitions the sequence into chunks, processes each chunk sequentially, and introduces an additional alignment stage to reconcile all chunks. 
FastVGGT~\cite{shen2025fastvggt} observes that both camera tokens and patch tokens exhibit a high degree of similarity in every global attention layer and, based on this observation, applies token merging~\cite{bolya2022token,haurum2023tokens,renggli2022learning,zeng2022not} to reduce the token sequence length when computing global attention. 
FasterVGGT~\cite{wang2025faster} conducts a more extensive analysis of VGGT than FastVGGT, yielding the following observations: 
(i) the global attention matrix is often sparse, 
(ii) its largest entries appear to correspond to geometrically meaningful correspondences, and 
(iii) intermediate global attention layers contribute more to performance than earlier or later ones. 
Given these insights, the acceleration strategy adopts SpargeAttention~\cite{dao2021pixelated,zhangspargeattention} to speed up global attention. 
Although promising, this choice appears only weakly related to the reported empirical findings.
In contrast, we thoroughly analyze VGGT and design our method fully guided by our findings.

%% file: section/3_analysis.tex
\begin{figure*}
	\centering
	\includegraphics[width=\linewidth]{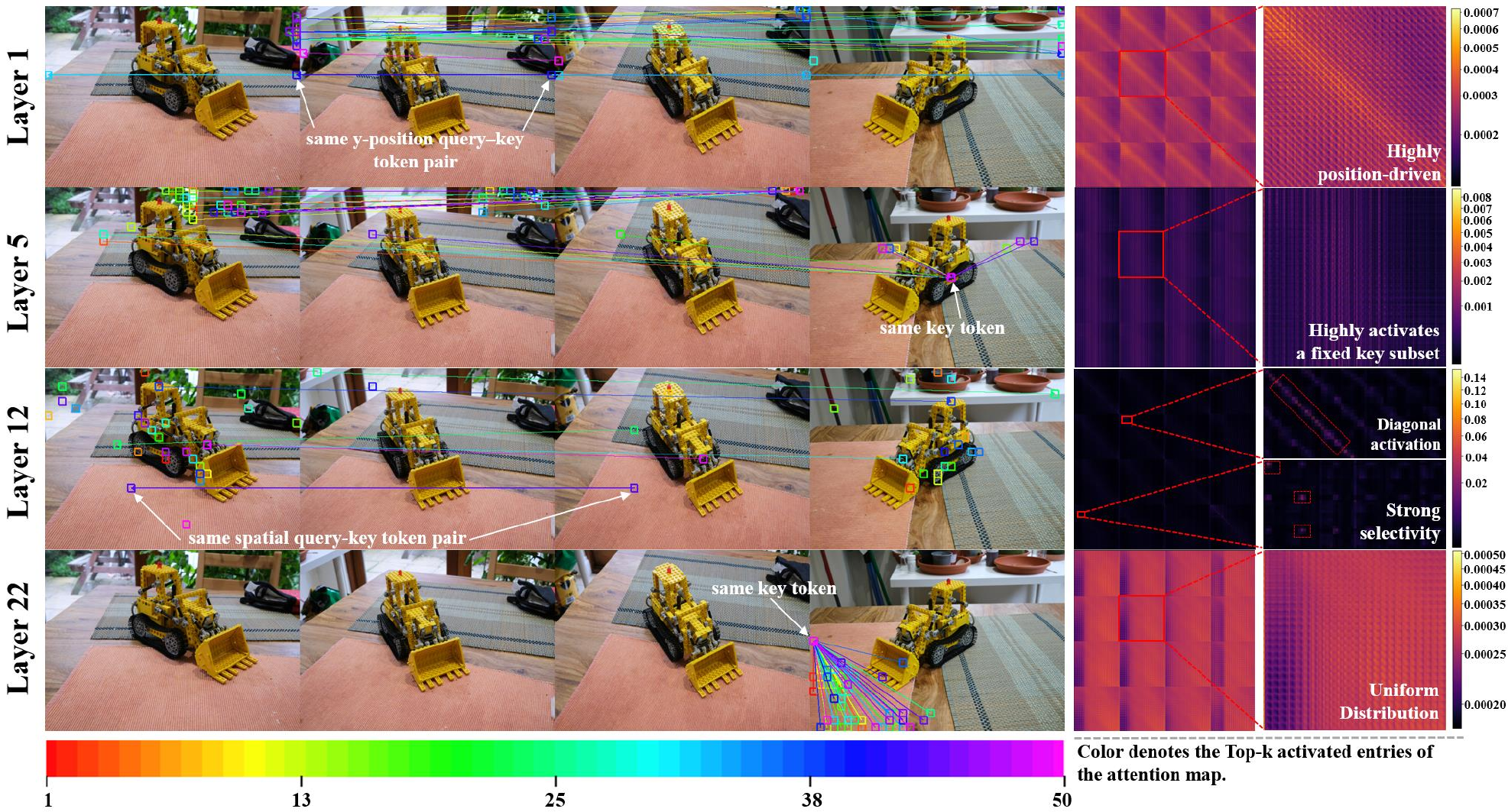}
    \caption{\textbf{Visualization of global attention across layers in VGGT.}
    Each row shows one global attention layer (Layers 1, 5, 12, 22; global layers indexed 0–23).
    \textbf{Left}: four input views with the Top-$k$ ($k=50$) attention entries overlaid; each arrow \emph{starts} at the query patch and \emph{ends} at the selected key patch; highlighted patches without arrows denote \emph{self} entries (query=key).
    \textbf{Right}: the multi-head attention matrix (averaged over heads) after softmax, with a zoomed-in region.
    For detailed analysis, please refer to Sec.~\ref{subsec:analy}.
    }
	\label{fig:vis_attn_map}
\end{figure*}

\section{Analyzing VGGT Global Attention}
\label{sec:analy}
We first outline the VGGT architecture. 
Given multiple input images, VGGT uses a pretrained DINOv2~\cite{oquab2023dinov2} encoder to extract patch features. 
For each frame, VGGT appends one learnable camera token and four learnable register tokens~\cite{darcet2023vision}. 
Since VGGT treats the first frame as the reference view, the camera and register tokens of the reference frame differ from those of the remaining frames. 
All frame tokens are then fed into the aggregator, which consists of 48 transformer blocks alternating between global and frame attention. 
After aggregation, all register tokens are removed, and the remaining tokens are sent to the camera head for pose estimation and to the DPT~\cite{ranftl2021vision} head for dense depth and point map prediction.

\subsection{Analyzing Global Attention Layers}
\label{subsec:analy}
We analyze four representative global layers (1, 5, 12, 22).
Attention is averaged over heads and restricted to patch tokens (special tokens are omitted for clarity).
The Top-$k$ ($k=50$) attention pairs with the highest weights are selected from the full attention matrix.
Fig.~\ref{fig:vis_attn_map} overlays arrows from query to key and shows the corresponding attention matrices.

\noindent\textbf{Early Global Attention Layers.}
In VGGT, the DINOv2 encoder is fine-tuned during training rather than kept frozen. However, such fine-tuning may not substantially reshape the underlying feature space, so the extracted features may still not contain sufficiently rich 3D information, while the overall architecture and training objectives target 3D reasoning. This raises a key question: \emph{Do the early global attention layers contribute to multi-view correlations?}
As shown in Fig.~\ref{fig:vis_attn_map}, the top activated entries in the attention maps of the early global attention layers (e.g., Layers~1 and~5) are much weaker than those in the middle layers (e.g., Layer~12), indicating a more uniform attention distribution. 
Moreover, examining the tokens corresponding to the top activated entries reveals two typical patterns. 
First, in the first one or two layers (e.g., Layer~1), the strongest matches almost always link tokens sharing the same \(y\)-coordinate, suggesting that attention at this stage is \textit{dominated by positional embeddings rather than image content}.
Second, in the other early layers (e.g., Layer~5), the top entries frequently attend to a small, fixed set of key tokens. 
To test whether these keys encode stable spatial semantics, we rotate all input images by $180^\circ$ (equivalent to rotating the camera while keeping the scene unchanged).
After rotation, the highly attended keys shift to different tokens, showing that these hubs do not correspond to any view-invariant 3D structure (see \textit{Supplementary Materials}). 
Taken together, these observations indicate that \emph{early global attention weights are not driven by spatial correspondence} and thus are unlikely to contribute to cross-view correlation. 
This conclusion is further supported by two ablations: replacing early Keys/Values with a mean token, or replacing early global attention with frame attention, both of which preserve accuracy (Sec.~\ref{subsec:ablation}).

\noindent\textbf{Middle Global Attention Layers.}
As shown in Fig.~\ref{fig:vis_attn_map}, the middle global attention layers (e.g., Layer~12) exhibit a qualitatively different behavior.
The attention maps become much sparser, and the top activated values increase sharply, indicating stronger selectivity. 
More importantly, the tokens associated with the highest activations fall almost exclusively into two categories: (i) the query and key correspond to the same token (self patch), and (ii) the two tokens correspond to the same spatial position across different views. 
Both patterns are directly related to 3D consistency, suggesting that \textit{the middle global layers are where the model truly begins to form multi-view correlations}. 
Inspired by these two attention patterns, where attention assigns high weights to patch tokens at the same spatial location, we propose the following hypothesis. 
From a point-cloud perspective, alternating global and frame attention acts as an iterative refinement process. 
\emph{Global attention performs alignment by linking spatially corresponding patches across views, whereas frame attention refines intra-view structure.}
To test this hypothesis, we perform a restricted global attention experiment (which also serves as the basis of our acceleration strategy) in which only a subset of tokens are used as Keys and Values, leaving the remaining tokens invisible to others in global attention during the forward pass. 
The final performance remains nearly unchanged, which partially supports the view that global attention primarily serves alignment. 
Detailed results and further analyses regarding the subsampling factor are provided in Sec.~\ref{subsec:ablation}.

\noindent\textbf{Last Global Attention Layers.}
From Fig.~\ref{fig:vis_attn_map}, we observe that the attention maps in the last several global attention layers (e.g., Layer~22) resemble those in the early layers. 
Their top attention values are noticeably smaller than those of the middle layers and exhibit a more uniform distribution. 
In addition, the highest activated attention entries also tend to concentrate on a small subset of key tokens.
Based on our hypothesis for the middle global attention layers, we infer that, from a point cloud perspective, the global point clouds are already nearly aligned in these layers. 
This implies that the global attention at this stage makes only minor adjustments to the alignment, and thus contributes little to multi-view consistency. 
As presented in Sec.~\ref{subsec:ablation}, directly replacing these final global attention layers with frame attention results in only a slight performance degradation. 
However, unlike the early global attention layers, the last global attention layers still make a minor yet non-negligible contribution to maintaining multi-view consistency, even though they are not the primary source of alignment.

\subsection{Summary}
In summary, our observations on VGGT global attention are as follows: 
In the alternating architecture of global and frame attention, global attention mainly discovers multi-view correspondences by aligning spatially corresponding tokens, while frame attention refines the structure within each view. 
Following this perspective, the early global layers are ineffective because the features at this depth contain little 3D signal, making meaningful correspondence impossible.  
In the middle layers, once frame attention has formed stable per-view structure, global attention can effectively perform cross-view alignment—this is where multi-view correlations truly emerge. 
In the last layers, the representations are already well aligned, so global attention provides only minor refinements rather than establishing new correspondences. 
A similar analysis for \(\pi^3\) is provided in the \textit{Supplementary Materials}.

%% file: section/4_method.tex
\begin{figure}
	\centering
	\includegraphics[width=\columnwidth]{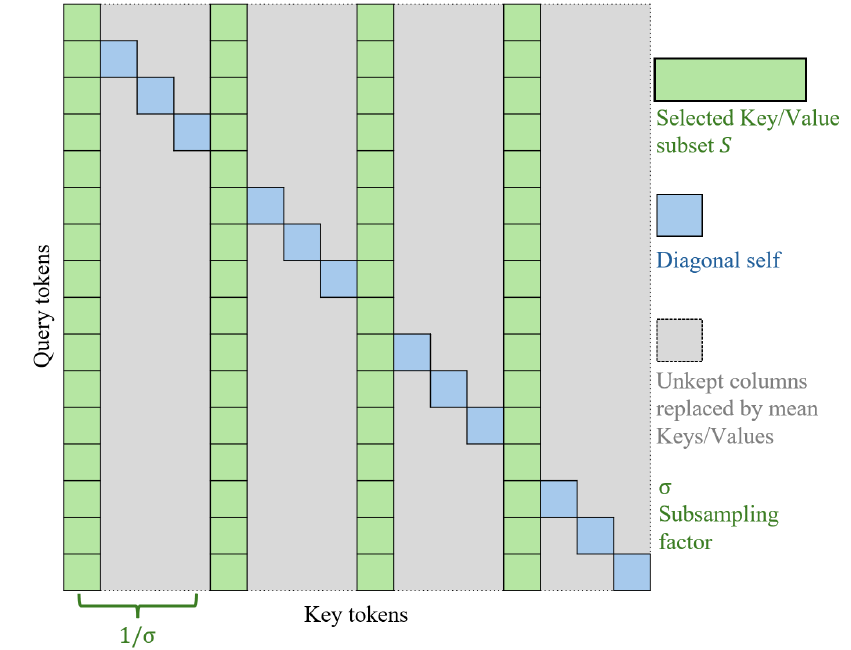}
    \caption{\textbf{Subsampling Global Attention (SGA).}
    Rows are queries and columns are keys.
    For patch tokens, a fixed per-frame subset $S$ forms the keys/values (one patch per $s_h\times s_w$ window, density $1/\sigma$).
    We preserve the diagonal self-interaction, approximate all dropped columns by a single mean component, and use a shared softmax normalization across all parts.
    See Sec.~\ref{subsec:dga} for details.}
	\label{fig:subsample_ga}
\end{figure}

\section{Methodology}
\label{sec:method}
Guided by the layer-wise analysis in Sec.~\ref{sec:analy}, our acceleration is \textbf{training-free} and has two steps:
(1) \emph{Global-to-Frame} conversion for early global layers; and
(2) \emph{Subsampling Global Attention (SGA)} on the remaining global layers by subsampling K/V over patch tokens while keeping all queries and special tokens.

\subsection{Notation}
In VGGT, given \(N\) input images, the \(i\)-th image produces a grid of patch tokens of size \(n_h \times n_w\) together with several special tokens (one camera token and four register tokens). 
Let \(T = n_h n_w\) denote the number of patch tokens per image, and \(L = T + 5\) the total number of tokens per frame. 
For clarity, \(N\) denotes the number of input frames, \(L\) the number of tokens per frame, and \(C\) the embedding dimension. 
Unless otherwise stated, attention operates on tensors of shape \((B, N, L, C)\), where \(B\) is the batch size.

\subsection{Global-to-Frame Attention Conversion}
\label{subsec:g2f}
Based on the layer-wise analysis in Sec.~\ref{sec:analy}, the early global attention layers do not contribute to multi-view correlations. 
We therefore convert these early global layers into frame attention to reduce computation and memory while preserving accuracy. 
In the original VGGT, tokens after a frame attention block are laid out with shape \((B N,\, L,\, C)\). Before a global attention block, the layout is rearranged to \((B,\, N L,\, C)\) so that attention is computed across all frames jointly. 
The same rearrangement is applied to positional embeddings. 
To convert a chosen global attention block into frame attention, we simply skip this rearrangement and keep the per-frame layout, applying attention independently to each frame while leaving all other components and parameters unchanged. 
Special tokens are kept per frame as in the original design.
This conversion reduces the cost of the affected block from $O((NL)^2)$ to $O(NL^2)$.

The last global attention layers still provide a minor benefit, so we keep them unchanged in order to preserve the capacity of the model.
Let the global attention blocks be indexed from \(0\) in order of depth. 
We introduce a split index \(t_{\mathrm{early}}\) that separates early from middle layers. 
Blocks with index \(< t_{\mathrm{early}}\) are converted to frame attention, and the remaining global attention blocks are kept. 
Unless otherwise noted, we set \(t_{\mathrm{early}} = 9\) for VGGT (thus converting indices \(0\) through \(8\)) and \(t_{\mathrm{early}} = 10\) for \(\pi^3\). 
Further details and ablations are provided in Sec.~\ref{subsec:ablation}.

\subsection{Subsampling Global Attention}
\label{subsec:dga}
After converting the early global attention layers to frame attention, the remaining global attention layers contribute to multi-view correlations to varying degrees (see Sec.~\ref{sec:analy}). 
Our analysis in Sec.~\ref{sec:analy} shows that global attention primarily performs alignment by linking spatially sparse corresponding patches across views, rather than requiring dense token-to-token matching. 
This implies that the full set of patch tokens is unnecessary for the alignment stage.
To exploit this redundancy, we introduce \emph{subsampling global attention}, where only a compact and spatially uniform subset of patch tokens is used as Keys and Values, while all Queries and all special tokens are preserved. \textbf{The reason for not subsampling Queries is that subsampling them reduces the set of tokens that receive cross-view updates, collapsing token diversity and harming dense 3D prediction. More detailed analysis and related ablation studies are provided in the \textit{Supplementary Materials}.}

\noindent\textbf{Subsampling Strategy.}
We introduce a total subsampling factor \(\sigma \in \mathbb{N}\) to control how many tokens are preserved. 
The key design question is how to choose which tokens to keep. 
Intuitively, from a point cloud perspective, when aligning two point clouds under subsampling, a reasonable strategy is to keep one point cloud intact while uniformly subsampling the other. 
Inspired by this intuition, we generalize the concept of point cloud alignment to feature maps by treating each patch token as a pseudo point in a spatial grid. 
We then perform grid-based subsampling on this patch grid, which serves as a 2D analogue of uniform point sampling in 3D space.
Specifically, we factorize \(\sigma = s_h s_w\), where \(s_h\) and \(s_w\) denote the strides along the height and width of the patch grid, respectively. 
In this work, we set $\sigma$ to map to $(s_h, s_w)$ as follows: $\sigma=2 \Rightarrow (s_h,s_w)=(1,2)$, $\sigma=4 \Rightarrow (s_h,s_w)=(2,2)$, $\sigma=6 \Rightarrow (s_h,s_w)=(2,3)$, and $\sigma=9 \Rightarrow (s_h,s_w)=(3,3)$.
For each frame with a patch grid of size \(n_h \times n_w\), we retain the first patch token within each \(s_h \times s_w\) window. 
For VGGT, the first frame is kept uncompressed as it serves as the reference view, while for \(\pi^3\), all frames are uniformly subsampled. 
We then compute attention using the reduced Key–Value set as described above. 
To verify the effectiveness of this design, we also evaluated alternative token-selection schemes, including random grid sampling and SIFT-based keypoint selection~\cite{lowe2004distinctive}.
As detailed in the \textit{Supplementary Materials}, our fixed grid-based approach consistently achieves both the highest accuracy and the fastest inference speed among all tested strategies.
By applying this strategy, the global attention computation becomes approximately \(\sigma\) times faster while maintaining alignment quality.

\noindent\textbf{Enhanced Subsampling with Diagonal Preservation.}
Although the basic subsampling strategy already achieves state-of-the-art performance, we conducted extensive exploration and refinement to further enhance its effectiveness. Motivated by our analysis in Sec.~\ref{sec:analy}, where highly activated attention entries are typically either diagonal or correspond to cross-view matches, we explicitly preserve the self-attention term (diagonal entry) for each token to maintain local feature coherence, and approximate all dropped columns with a single mean Key–Value pair that captures the aggregated global response. The attention is computed over three disjoint components: (i) the retained subset \(S\), (ii) the diagonal (self term), and (iii) a mean component representing all dropped patches. These components share a single softmax normalization so that their attention weights are jointly normalized without redundancy. Ablation results show that this refinement yields nearly identical performance to the base strategy under sparse input settings (\textit{e.g.}, 10 frames), while providing slight improvements when the inputs become dense (\textit{e.g.}, 300 frames). Detailed results can be found in the \textit{Supplementary Materials}. Most importantly, this additional component introduces only \(O(N)\) computational overhead and does not affect the overall acceleration.

%% file: section/5_experiments.tex
\begin{table*}[t]
  \centering
  \caption{
    Camera pose estimation on RealEstate10K~\cite{zhou2018stereo} under the sparse setting (10 frames).
    Best and second best are highlighted within each baseline block excluding the baseline row.
  }
  \label{tab:sparse_ang}
  \resizebox{\textwidth}{!}{%
  \begin{tabular}{llccccccccccc}
    \hline
    Baseline & Method &
    RRA@5 $\uparrow$ & RTA@5 $\uparrow$ & AUC@5 $\uparrow$ &
    RRA@15 $\uparrow$ & RTA@15 $\uparrow$ & AUC@15 $\uparrow$ &
    RRA@30 $\uparrow$ & RTA@30 $\uparrow$ & AUC@30 $\uparrow$ &
    Time (s) $\downarrow$ & FLOPs (T) $\downarrow$ \\
    \hline
    \multirow{5}{*}{\(\pi^3\)}
      & \(\pi^3\)              & 98.777 & 83.882 & 67.186 & 99.817 & 94.147 & 83.288 & 99.954 & 96.593 & 89.500 & 0.220 & 10.449 \\
      & Faster\_\(\pi^3\)\_25  & \textbf{98.782} & \underline{80.879} & \underline{62.950} & \underline{99.803} & 92.716 & \underline{80.545} & 99.904 & 95.856 & \underline{87.639} & 0.237 & 9.998 \\
      & Faster\_\(\pi^3\)\_75  & 98.335 & 65.744 & 42.887 & \textbf{99.844} & 86.044 & 67.530 & \textbf{99.997} & 92.675 & 78.913 & 0.223 & 8.870 \\
      & A\_\(\pi^3\)(2)        & \underline{98.692} & \textbf{82.905} & \textbf{64.832} & \underline{99.803} & \textbf{94.060} & \textbf{82.306} & \underline{99.926} & \textbf{96.678} & \textbf{89.008} & \underline{0.207} & \underline{8.837} \\
      & A\_\(\pi^3\)(4)        & 98.559 & 79.375 & 58.703 & 99.697 & \underline{93.511} & 79.350 & 99.836 & \underline{96.618} & 87.434 & \textbf{0.203} & \textbf{8.594} \\
    \hline
    \multirow{6}{*}{VGGT}
      & VGGT                   & 98.545 & 81.471 & 63.176 & 99.913 & 93.342 & 81.103 & 100.000 & 96.189 & 88.130 & 0.307 & 12.412 \\
      & FastVGGT               & 98.168 & 76.249 & 53.506 & \underline{99.896} & 91.832 & 76.167 & \textbf{100.000} & 95.687 & 85.236 & 0.735 & \underline{10.243} \\
      & FasterVGGT\_25         & \textbf{98.550} & 77.939 & 58.934 & \textbf{99.913} & 91.870 & 78.179 & \textbf{100.000} & 95.501 & 86.212 & 0.326 & 11.812 \\
      & FasterVGGT\_75         & 97.546 & 61.540 & 38.794 & 99.495 & 84.242 & 64.586 & 99.607 & 91.351 & 76.651 & 0.307 & 10.309 \\
      & AVGGT(2)               & \underline{98.501} & \textbf{80.775} & \textbf{61.959} & 99.872 & \textbf{93.118} & \textbf{80.443} & 99.956 & \textbf{96.216} & \textbf{87.758} & \underline{0.298} & 10.580 \\
      & AVGGT(4)               & 98.444 & \underline{79.337} & \underline{59.583} & 99.844 & \underline{92.722} & \underline{79.188} & 99.932 & \underline{96.118} & \underline{87.045} & \textbf{0.291} & \textbf{10.170} \\
    \hline
  \end{tabular}
  }
\end{table*}

\begin{table}[t]
  \centering
  \caption{
    Camera pose estimation on TUM-dynamics~\cite{sturm2012benchmark} under the sparse setting (90 frames).
    Best and second best are highlighted within each baseline block, excluding the baseline row.
  }
  \label{tab:sparse_dis}
  \resizebox{\columnwidth}{!}{%
  \begin{tabular}{llccccc}
    \hline
    Baseline & Method & ATE $\downarrow$ & RPE trans $\downarrow$ & RPE rot $\downarrow$ & Time (s) $\downarrow$ & FLOPs (T) $\downarrow$ \\
    \hline
    \multirow{5}{*}{\(\pi^3\)} 
      & \(\pi^3\)                 & 0.014 & 0.009 & 0.307 & 5.768 & 430.038 \\
      & Faster\_\(\pi^3\)\_25     & \textbf{0.014} & \textbf{0.009} & \textbf{0.307} & 5.242 & 365.321 \\
      & Faster\_\(\pi^3\)\_75     & 0.017 & 0.011 & 0.324 & 3.348 & 203.528 \\
      & A\_\(\pi^3\)(2)           & \textbf{0.014} & \textbf{0.009} & \underline{0.309} & \underline{2.921} & \underline{183.589} \\
      & A\_\(\pi^3\)(4)           & 0.015 & \textbf{0.009} & 0.311 & \textbf{2.508} & \textbf{146.842} \\
    \hline
    \multirow{6}{*}{VGGT} 
      & VGGT                      & 0.012 & 0.010 & 0.309 & 7.924 & 508.650 \\
      & FastVGGT                  & 0.013 & 0.011 & 0.321 & 4.733 & \underline{210.020} \\
      & FasterVGGT\_25            & 0.012 & \textbf{0.010} & 0.312 & 7.303 & 451.390 \\
      & FasterVGGT\_75            & 0.017 & 0.012 & 0.346 & 4.722 & 240.720 \\  
      & AVGGT(2)                  & \textbf{0.012} & \textbf{0.010} & \underline{0.310} & \underline{4.519} & 253.330 \\
      & AVGGT(4)                  & \textbf{0.012} & \textbf{0.010} & \textbf{0.309} & \textbf{3.761} & \textbf{186.790} \\
    \hline
  \end{tabular}
  }
\end{table}

\begin{table}[t]
  \centering
  \caption{%
    Point map estimation on DTU~\cite{jensen2014large} under the sparse setting (10 frames).
    Best and second best are highlighted within each baseline block, excluding the baseline row.
  }
  \label{tab:sparse_point}
  \resizebox{\columnwidth}{!}{%
  \begin{tabular}{llcccccccc}
    \hline
    \multirow{2}{*}{Baseline} & \multirow{2}{*}{Method} &
    \multicolumn{2}{c}{Acc. $\downarrow$} &
    \multicolumn{2}{c}{Comp. $\downarrow$} &
    \multicolumn{2}{c}{NC. $\uparrow$} &
    \multirow{2}{*}{Time (s) $\downarrow$} &
    \multirow{2}{*}{FLOPs (T) $\downarrow$} \\
    \cline{3-4}\cline{5-6}\cline{7-8}
     &  & Mean & Med. & Mean & Med. & Mean & Med. &  &  \\
    \hline
    \multirow{5}{*}{\(\pi^3\)}
      & \(\pi^3\)                 & 1.152 & 0.623 & 1.801 & 0.631 & 0.668 & 0.754 & 0.380 & 15.220 \\
      & Faster\_\(\pi^3\)\_25     & \textbf{1.210} & \textbf{0.645} & \textbf{1.844} & \textbf{0.624} & 0.668 & 0.755 & 0.406 & 14.421 \\
      & Faster\_\(\pi^3\)\_75     & 2.451 & 1.283 & 2.236 & 0.795 & \textbf{0.680} & \textbf{0.768} & 0.379 & 12.424 \\
      & A\_\(\pi^3\)(2)           & \underline{1.261} & \underline{0.669} & \underline{1.870} & \underline{0.628} & 0.667 & 0.753 & \underline{0.351} & \underline{12.363} \\
      & A\_\(\pi^3\)(4)           & 1.598 & 0.802 & 2.060 & 0.691 & \underline{0.670} & \underline{0.757} & \textbf{0.320} & \textbf{11.909} \\
    \hline
    \multirow{6}{*}{VGGT}
      & VGGT                      & 1.185 & 0.714 & 2.224 & 1.307 & 0.694 & 0.779 & 0.466 & 18.113 \\
      & FastVGGT                  & 1.381 & 0.784 & 2.532 & 1.838 & 0.576 & 0.610 & 0.805 & \underline{14.167} \\
      & FasterVGGT\_25            & 1.194 & 0.725 & \underline{2.168} & 1.236 & \underline{0.691} & \textbf{0.775} & 0.555 & 17.047 \\
      & FasterVGGT\_75            & 1.445 & 0.830 & \textbf{2.131} & \textbf{1.117} & 0.674 & 0.761 & 0.477 & 14.384 \\
      & AVGGT(2)                  & \textbf{1.177} & \textbf{0.714} & 2.180 & 1.263 & \textbf{0.692} & \textbf{0.775} & \underline{0.458} & 14.865 \\
      & AVGGT(4)                  & \underline{1.193} & \underline{0.715} & \underline{2.168} & \underline{1.229} & 0.689 & 0.773 & \textbf{0.428} & \textbf{14.099} \\
    \hline
  \end{tabular}
  }
\end{table}

\begin{table*}[t]
  \centering
  \caption{
    Camera pose and point map estimation on 7-Scenes~\cite{shotton2013scene} under the dense setting (333 frames).
  }
  \label{tab:dense_all}
  \resizebox{\textwidth}{!}{%
  \begin{tabular}{ll|ccccccccc|cccccc|cc}
    \hline
    \multirow{2}{*}{Baseline} & \multirow{2}{*}{Method} &
    \multicolumn{9}{c|}{\textbf{Camera Pose Estimation}} &
    \multicolumn{6}{c|}{\textbf{Point Map Estimation}} &
    \multirow{2}{*}{Time (s) $\downarrow$} &
    \multirow{2}{*}{FLOPs (P) $\downarrow$} \\
    \cline{3-17}
     &  &
    R@5 $\uparrow$ & T@5 $\uparrow$& AUC@5 $\uparrow$&
    R@15 $\uparrow$& T@15 $\uparrow$& AUC@15 $\uparrow$&
    R@30 $\uparrow$& T@30 $\uparrow$& AUC@30 $\uparrow$&
    Acc.M $\downarrow$& Acc.Md $\downarrow$&
    Comp.M $\downarrow$& Comp.Md $\downarrow$&
    NC.M $\uparrow$& NC.Md $\uparrow$&  &  \\
    \hline
    \multirow{5}{*}{$\pi^3$}
      & $\pi^3$                      & 78.122 & 68.632 & 27.450 & 99.872 & 93.254 & 65.289 & 100.000 & 97.443 & 80.674 & 0.019 & 0.003 & 0.035 & 0.016 & 0.536 & 0.552 & 59.175 & 4.832 \\
      & Faster\_$\pi^3$\_25          & \textbf{99.374} & 50.363 & 25.537 & \textbf{100.000} & 85.553 & 58.620 & \textbf{100.000} & 94.915 & 75.054 & \textbf{0.019} & \textbf{0.003} & 0.034 & 0.015 & \underline{0.536} & \underline{0.552} & 50.850 & 3.896 \\
      & Faster\_$\pi^3$\_75          & \underline{98.588} & 38.761 & 19.637 & \textbf{100.000} & 80.387 & 51.839 & \textbf{100.000} & 93.019 & 69.916 & 0.021 & 0.004 & \textbf{0.025} & \textbf{0.004} & \textbf{0.543} & \textbf{0.562} & 23.146 & 1.579 \\
      & A\_$\pi^3$(2)                & 78.295 & \textbf{69.567} & \textbf{27.868} & 99.954 & \textbf{93.027} & \textbf{65.454} & \textbf{100.000} & \textbf{97.455} & \textbf{80.722} & \textbf{0.019} & \textbf{0.003} & \underline{0.033} & \underline{0.011} & \underline{0.536} & \underline{0.552} & \underline{17.909} & \underline{1.404} \\
      & A\_$\pi^3$(4)                & 75.607 & \underline{68.416} & \underline{26.978} & 99.962 & \underline{92.688} & \underline{64.683} & \textbf{100.000} & \underline{97.423} & \underline{80.270} & \textbf{0.019} & \textbf{0.003} & \underline{0.033} & \underline{0.011} & \underline{0.536} & \underline{0.552} & \textbf{12.608} & \textbf{0.898} \\
    \hline
    \multirow{6}{*}{VGGT}
      & VGGT                         & 73.262 & 59.605 & 24.943 & 99.939 & 89.854 & 61.118 & 100.000 & 96.576 & 77.755 & 0.054 & 0.029 & 0.148 & 0.092 & 0.525 & 0.536 & 79.945 & 6.369 \\
      & FastVGGT                     & 70.366 & 58.266 & 22.477 & 99.995 & 89.401 & 59.789 & \textbf{100.000} & 96.288 & 76.898 & \underline{0.054} & \underline{0.029} & \underline{0.136} & \underline{0.082} & 0.521 & 0.529 & \underline{28.025} & \underline{1.702} \\
      & FasterVGGT\_25               & \textbf{99.052} & 44.471 & 23.354 & \textbf{100.000} & 83.838 & 55.378 & \textbf{100.000} & 95.460 & 73.232 & 0.056 & 0.030 & 0.149 & 0.094 & 0.525 & 0.535 & 69.995 & 5.121 \\
      & FasterVGGT\_75               & \underline{98.931} & 38.236 & 18.422 & \textbf{100.000} & 81.154 & 51.322 & \textbf{100.000} & 93.402 & 70.074 & \textbf{0.053} & \textbf{0.027} & 0.147 & 0.090 & \textbf{0.528} & \textbf{0.541} & 32.492 & 2.031 \\
      & AVGGT(2)                     & 75.814 & \underline{60.357} & \underline{26.061} & 99.997 & \textbf{89.744} & \underline{61.951} & \textbf{100.000} & \textbf{96.482} & \underline{78.113} & \underline{0.054} & 0.030 & 0.142 & 0.088 & 0.527 & 0.539 & 30.430 & 2.354 \\
      & AVGGT(4)                     & 77.447 & \textbf{61.143} & \textbf{26.405} & 99.983 & \underline{89.691} & \textbf{62.369} & \textbf{100.000} & \underline{96.480} & \textbf{78.290} & \underline{0.054} & 0.031 & \textbf{0.131} & \textbf{0.078} & \textbf{0.528} & \textbf{0.541} & \textbf{20.642} & \textbf{1.408} \\
    \hline
  \end{tabular}
  }%
\end{table*}

\begin{table}[t]
  \centering
  \caption{
    Camera pose estimation on 7-Scenes~\cite{shotton2013scene} under the extremely dense setting (800 frames). \textit{OOM} denotes out-of-memory.
  }
  \label{tab:extreme_dense}
  \resizebox{\columnwidth}{!}{%
  \begin{tabular}{llccccc}
    \hline
    Baseline & Method & AUC@5 $\uparrow$ & AUC@15 $\uparrow$ & AUC@30 $\uparrow$ & Time (s) $\downarrow$ & FLOPs (P) $\downarrow$ \\
    \hline
    \multirow{7}{*}{$\pi^3$}
      & $\pi^3$                 & 27.429 & 65.195 & 80.572 & 298.477 & 26.465 \\
      & Faster\_$\pi^3$\_25     & \textit{OOM} & \textit{OOM} & \textit{OOM} & \textit{OOM} & \textit{OOM} \\
      & Faster\_$\pi^3$\_75     & \textit{OOM} & \textit{OOM} & \textit{OOM} & \textit{OOM} & \textit{OOM} \\
      & A\_\(\pi^3\)(2)         & \textbf{27.763} & \textbf{65.303} & \textbf{80.590} & 71.107 & 6.708 \\
      & A\_\(\pi^3\)(4)         & \underline{26.974} & \underline{64.594} & \underline{80.169} & 44.709 & 3.805 \\
      & A\_\(\pi^3\)(6)         & 26.335 & 64.057 & 79.748 & \underline{35.776} & \underline{2.888} \\
      & A\_\(\pi^3\)(9)         & 26.579 & 64.300 & 79.409 & \textbf{30.273} & \textbf{2.267} \\
    \hline
    \multirow{7}{*}{VGGT}
      & VGGT                    & 23.551 & 57.815 & 74.161 & 397.133 & 35.113 \\
      & FastVGGT                & 22.116 & 59.178 & 76.461 & 103.184 & 8.312 \\
      & FasterVGGT\_25          & \textit{OOM} & \textit{OOM} & \textit{OOM} & \textit{OOM} & \textit{OOM} \\
      & FasterVGGT\_75          & \textit{OOM} & \textit{OOM} & \textit{OOM} & \textit{OOM} & \textit{OOM} \\
      & AVGGT(2)                & 24.716 & 59.155 & 75.064 & 126.913 & 12.043 \\
      & AVGGT(4)                & \textbf{25.508} & 61.188 & 77.368 & 76.500 & 6.605 \\
      & AVGGT(6)                & \underline{25.354} & \textbf{61.349} & \textbf{77.461} & \underline{59.963} & \underline{4.888} \\
      & AVGGT(9)                & 24.896 & \underline{61.334} & \underline{77.382} & \textbf{50.034} & \textbf{3.825} \\
    \hline
  \end{tabular}
  }
\end{table}

\section{Experiments}
\label{label:exp}
\subsection{Experimental Setup}
We evaluate our acceleration strategy on VGGT~\cite{wang2025vggt} and \(\pi^3\)~\cite{wang2025pi}, denoted as AVGGT and A\_\(\pi^3\), respectively. 
Since our method supports different subsampling factors \(\sigma\) for global attention, we use AVGGT(2) to indicate \(2\times\) subsampling, and adopt the same notation for A\_\(\pi^3\). 
As FasterVGGT~\cite{wang2025faster} also supports accelerating \(\pi^3\), we include this variant as Faster\_\(\pi^3\). 
FasterVGGT exposes two tunable parameters: the CDF threshold \(\tau\) and the sparse ratio \(\rho\). 
We report two official configurations, \((\rho,\tau)=(0.25,0.9)\) and \((\rho,\tau)=(0.75,0.4)\), denoted as FasterVGGT\_25 and FasterVGGT\_75, and use the same naming convention for Faster\_\(\pi^3\).
To assess the effectiveness under different view densities, we consider both sparse and dense settings. 
In the sparse setting, we use RealEstate10K~\cite{zhou2018stereo} and TUM-dynamics~\cite{sturm2012benchmark} for camera pose estimation, and DTU~\cite{jensen2014large} for point-map estimation. 
In the dense setting, we use 7-Scenes~\cite{shotton2013scene} for both pose and point-map estimation.
In addition to the above results, further dataset results are available in the \textit{Supplementary Materials}.
For evaluation metrics, we follow the \(\pi^3\) protocol~\cite{wang2025pi}. 
For camera pose estimation, we report the Relative Rotation Accuracy (RRA) and Relative Translation Accuracy (RTA) at a given threshold (e.g., RRA@30 for 30 degrees). 
The Area Under the Curve (AUC) of the \(\min(\text{RRA}, \text{RTA})\) threshold curve serves as a unified metric. 
In addition, we report the Absolute Trajectory Error (ATE), Relative Pose Error for translation (RPE~trans.), and Relative Pose Error for rotation (RPE~rot.).
For point map estimation, we report Accuracy (Acc.), Completion (Comp.), and Normal Consistency (N.C.).
All experiments are conducted on NVIDIA A100 GPUs (80 GiB VRAM), with FlashAttention-2~\cite{dao2023flashattention} enabled during inference.

\subsection{Sparse Settings}
We follow the \(\pi^3\) evaluation protocol~\cite{wang2025pi}. Sequences in RealEstate10K and DTU contain 10 frames, whereas sequences in TUM-dynamics contain 90 frames.
Camera pose estimation results are reported in Table~\ref{tab:sparse_ang} and Table~\ref{tab:sparse_dis}. FasterVGGT is comparable to the baselines when \(\rho{=}0.25\), but accuracy drops clearly when \(\rho\) increases to \(0.75\). In contrast, our method remains stable when the subsampling factor changes from \(\sigma{=}2\) to \(\sigma{=}4\). 
FastVGGT stays close to the baselines on TUM-dynamics where each sequence is long, but degrades noticeably on RealEstate10K where sequences are short. 
Because both FasterVGGT and FastVGGT introduce extra computation, on RealEstate10K where each sequence has only 10 frames, they are even slower than the original models, whereas our method yields a small but consistent speedup under these settings. 
Point map estimation results on DTU are reported in Table~\ref{tab:sparse_point}, the trends mirror those for pose: FastVGGT drops on short sequences and FasterVGGT degrades when \(\rho\) increases to \(0.75\), while our method accelerates inference without loss of accuracy. 
Overall, our approach provides measurable speedup without affecting accuracy.

\subsection{Dense Settings}
In the 7-Scenes dataset, each scene contains 1000 frames. 
For the dense setting, we subsample with a stride of 3, yielding test sequences of 333 frames per scene. 
The results are shown in Table~\ref{tab:dense_all}. 
For camera pose estimation, while FasterVGGT often reports very high $\mathrm{RRA}$, our method achieves competitive, and often the best, overall accuracy across metrics and does so with lower runtime. 
For point-map estimation, we observe that both FasterVGGT and our method tend to improve as the sparse ratio (or our subsampling factor) increases; when using \(\pi^3\) as the baseline, FasterVGGT attains the best results, whereas with VGGT as the baseline our method performs best. 
Overall, accuracy differences under the dense setting are small, but our approach consistently delivers the largest speedup.

To further evaluate our acceleration performance under extremely dense views, we conduct an additional experiment on the 7-Scenes dataset by extending each sequence to 800 frames, measuring only inference time and pose estimation accuracy. 
As shown in Table~\ref{tab:extreme_dense}, for \(\pi^3\), even with a \(9\times\) subsampling factor, the results remain almost identical to the original while achieving nearly a \(10\times\) speedup. 
Similarly, for VGGT, under \(9\times\) subsampling, our strategy even outperforms the original method while achieving an \(8\times\) speedup.

\begin{figure}[t]
	\centering
	\includegraphics[width=\linewidth]{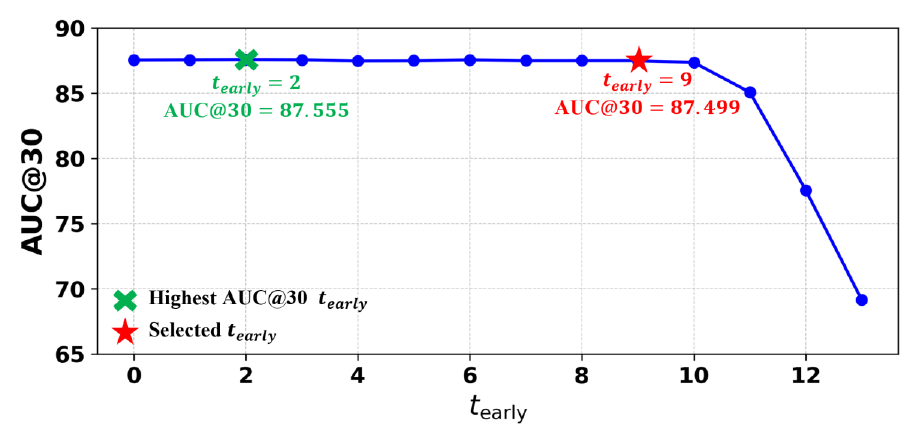}
	\caption{Ablation results on RealEstate10K for \(t_{early}\) choices.}
	\label{fig:abla_t_early}
\end{figure}

\begin{table}[t]
  \centering
  \caption{
    Ablation results on RealEstate10K~\cite{zhou2018stereo}.
  }
  \label{tab:aba_all}
  \resizebox{\columnwidth}{!}{%
  \begin{tabular}{lccc}
    \hline
    Method & AUC@5 $\uparrow$ & AUC@15 $\uparrow$ & AUC@30 $\uparrow$ \\
    \hline
      VGGT         & 63.176 & 81.103 & 88.130 \\
      VGGT(G2F)      & 62.826 & 80.882 & 87.981 \\
      VGGT(G2M)     & 61.788 & 80.064 & 87.425 \\
      AVGGT(2)$^-$    & 59.637 & 79.305 & 87.161 \\
      AVGGT(2)     & 61.959 & 80.443 & 87.758 \\
      AVGGT(4)     & 59.583 & 79.188 & 87.045 \\
      AVGGT(6)     & 57.177 & 77.757 & 86.113 \\
      AVGGT(9)     & 53.257 & 75.473 & 84.723 \\
    \hline
  \end{tabular}
  }
\end{table}

\subsection{Ablation Studies}
\label{subsec:ablation}

All ablation studies below are conducted on VGGT, with camera pose estimation evaluated on RealEstate10K~\cite{zhou2018stereo}, consistent with the sparse setting. 
For \(\pi^3\), the results follow the same trends, and detailed results are provided in the \textit{Supplementary Materials}. 

\noindent\textbf{Effect of the Early Global Attention Layers.}
To validate whether the early global attention layers contribute to building multi-view correlations, we design two ablation variants. 
The first variant, denoted as VGGT(G2F), directly replaces the early global attention layers (indices 0–8) with frame attention, as described in Sec.~\ref{subsec:g2f}. 
The second variant, denoted as VGGT(G2M), modifies the early global attention layers by keeping the Query tokens unchanged while replacing all Keys and Values with their mean representations across all tokens (including special tokens), resulting in a single Key–Value pair for the entire sequence.
As shown in Table~\ref{tab:aba_all}, VGGT(G2F) achieves nearly the same metrics as the original model, even though in this variant there is no information exchange across different views. 
This observation demonstrates that the early global attention layers are not essential for establishing multi-view correlations. 
Similarly, VGGT(G2M), which uses only one mean token for both Key and Value, yields comparable performance to the original model, indicating that these early layers do not perform meaningful selective attention.

\noindent\textbf{Effect of the \(t_{\text{early}}\) Choice.}
The choice of \(t_{\text{early}}\) can be roughly inferred from the global attention map analysis in Sec.~\ref{subsec:analy}, which indicates which global attention layers primarily contribute to building multi-view correlations. 
However, such analysis can only provide an approximate index. 
To preserve as much model capacity as possible while improving efficiency, we therefore conduct an ablation study to determine an appropriate \(t_{\text{early}}\) value. 
As shown in Fig.~\ref{fig:abla_t_early}, for VGGT, we set \(t_{\text{early}} = 9\) to achieve a balance between performance and speed.

\noindent\textbf{Effect of the Subsampling Factor \(\sigma\).}
We examine how the subsampling factor $\sigma$ affects performance. In our subsampled global attention, all Queries are retained, while a fixed subset of Keys and Values is used for global attention; this subset remains consistent across all global layers with subsampling. As shown in Table~\ref{tab:aba_all}, even with $\sigma=9$, the performance remains close to that of the original model. Moreover, in the dense case (Table~\ref{tab:extreme_dense}), where each scene contains 800 frames and strong cross-view overlap, $\sigma=9$ even outperforms the original. More importantly, these results further demonstrate that the global attention layers in the alternating architecture primarily serve alignment purposes: in the sparse setting, reducing the number of tokens used for alignment inevitably leads to performance degradation, whereas in the dense setting, the model becomes more robust to such reduction because of the redundancy of overlapping regions. 
In conclusion, our method maintains nearly original performance even under \(\sigma=9\), and the performance gap between sparse and dense settings further confirms that the global attention layers mainly function to establish cross-view alignment.

\noindent\textbf{Effect of the Last Global Attention Layers.}
To validate whether the last global attention layers are useful for building multi-view correlations, we conduct an ablation based on our AVGGT(2) variant. 
Specifically, we replace the last global attention layers (indices 20–23) with frame attention, denoted as AVGGT(2)\(^{-}\). 
As shown in Table~\ref{tab:aba_all}, this modification results in only a slight performance drop compared to AVGGT(2), indicating that the last global attention layers have only a minor impact on building multi-view correlations.

%% file: section/6_conclusion.tex
\section{Conclusion}
\label{sec:conclusion}
We present a training-free acceleration strategy for feed-forward 3D reconstruction, grounded in an in-depth analysis of VGGT. Our study clarifies the respective roles of global attention and frame attention, and shows how to preserve accuracy while reducing inference cost. Our analysis provides a clearer picture of how alternating attention operates in large 3D models, and we expect these observations to guide the design of future architectures and training objectives for general-purpose 3D perception.

\section{Acknowledgments}
This work was supported in part by the National Natural Science Foundation of China (Grant Nos. 62302295, 62595733, and 62561160155), the Shanghai Municipal Science and Technology Major Project (Grant No. 2021SHZDZX0102). This work was also supported by Ant Group Research Intern Program.

%% file: section/7_supp.tex
\clearpage
\setcounter{page}{1}
\maketitlesupplementary

\begin{figure*}
\centering
\resizebox{\textwidth}{!}{\includegraphics{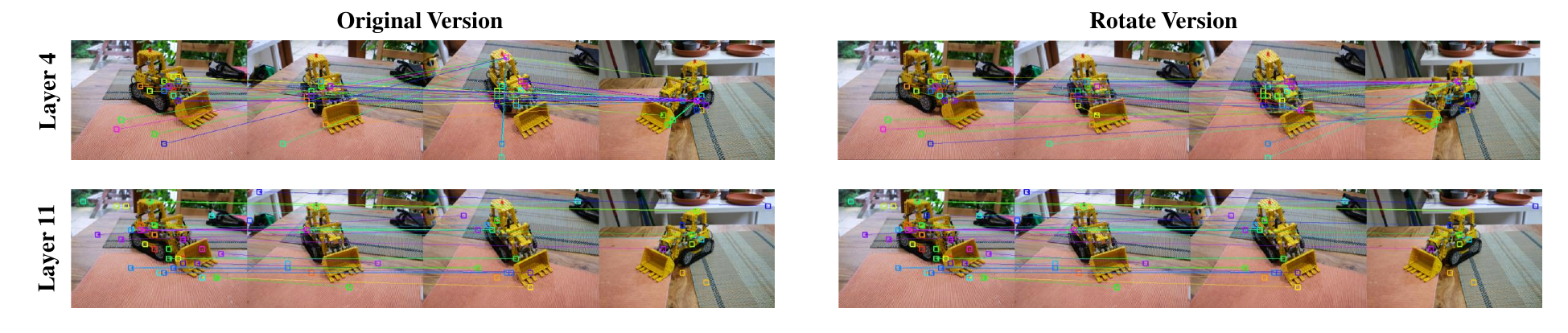}}
\caption{
Rotation test on highly activated key-token subsets in VGGT. 
All input views are rotated by $180^\circ$ and fed through VGGT, and the outputs are rotated back for visualization. 
For each layer, we first collect the top 1000 attention entries in both the original and rotated runs, then select the top 50 entries that share the same query patch (following the ranking in the original run). 
Arrows start at the query patch and end at the corresponding key patch.
}
\label{fig:rot_ablation}
\end{figure*}

\begin{table}[t]
  \centering
  \caption{
    Camera pose estimation on Sintel~\cite{bozic2021transformerfusion}.
    Best and second best are highlighted within each baseline block, excluding the baseline row.
  }
  \label{tab:sparse_dis_sintel}
  \resizebox{\columnwidth}{!}{%
  \begin{tabular}{llcccc}
    \toprule
    Baseline & Method & ATE $\downarrow$ & RPE trans $\downarrow$ & RPE rot $\downarrow$ & Time (s) $\downarrow$ \\
    \midrule
    \multirow{5}{*}{\(\pi^3\)} 
      & \(\pi^3\)                 & 0.073 & 0.038 & 0.288 & 0.960 \\
      & Faster\_\(\pi^3\)\_25     & \textbf{0.076} & \textbf{0.044} & \underline{0.318} & 1.020 \\
      & Faster\_\(\pi^3\)\_75     & 0.096 & 0.071 & 0.508 & 0.809 \\
      & A\_\(\pi^3\)(2)           & \underline{0.091} & \underline{0.046} & \textbf{0.304} & \underline{0.725} \\
      & A\_\(\pi^3\)(4)           & 0.127 & 0.061 & 0.345 & \textbf{0.687} \\
    \midrule
    \multirow{6}{*}{VGGT} 
      & VGGT                      & 0.169 & 0.061 & 0.476 & 1.348 \\
      & FastVGGT                  & \textbf{0.169} & 0.072 & 0.525 & 1.584 \\
      & FasterVGGT\_25            & \underline{0.171} & \textbf{0.063} & \underline{0.505} & 1.348 \\
      & FasterVGGT\_75            & 0.199 & 0.095 & 0.847 & 1.107 \\  
      & AVGGT(2)                  & 0.183 & \underline{0.068} & \textbf{0.497} & \underline{1.069} \\
      & AVGGT(4)                  & 0.199 & 0.087 & 0.539 & \textbf{1.006} \\
    \bottomrule
  \end{tabular}
  }
\end{table}

\begin{table}[t]
  \centering
  \caption{
     Camera pose estimation on ETH3D~\cite{schops2017multi}.
    Best and second best are highlighted within each baseline block, excluding the baseline row.
  }
  \label{tab:ETH3D}
  \resizebox{\columnwidth}{!}{%
  \begin{tabular}{llcccccccccc}
    \toprule
    Baseline & Method
    & Racc@5 $\uparrow$ & Tacc@5 $\uparrow$ & AUC@5 $\uparrow$
    & Racc@15 $\uparrow$ & Tacc@15 $\uparrow$ & AUC@15 $\uparrow$
    & Racc@30 $\uparrow$ & Tacc@30 $\uparrow$ & AUC@30 $\uparrow$
    & Time (s) $\downarrow$ \\
    \midrule
    \multirow{5}{*}{\(\pi^3\)}
      & \(\pi^3\)                 & 99.471 & 86.772 & 67.566 & 100.000 & 96.296 & 85.538 & 100.000 & 98.148 & 91.455 & 1.307 \\
      & Faster\_\(\pi^3\)\_25     & \textbf{99.735} & \textbf{80.159} & \textbf{60.106} & \textbf{100.000} & \underline{95.767} & \textbf{81.446} & \textbf{100.000} & 97.884 & \textbf{89.356} & 1.278 \\
      & Faster\_\(\pi^3\)\_75     & 96.032 & 67.989 & 42.222 & \textbf{100.000} & 91.799 & 70.176 & \textbf{100.000} & \textbf{98.677} & 83.254 & 0.991 \\
      & A\_\(\pi^3\)(2)           & \underline{98.413} & \underline{79.630} & \underline{52.857} & \textbf{100.000} & \textbf{96.561} & \underline{78.836} & \textbf{100.000} & \underline{98.413} & \underline{88.228} & \underline{0.955} \\
      & A\_\(\pi^3\)(4)           & 96.825 & 68.783 & 43.810 & \textbf{100.000} & 93.122 & 71.693 & \textbf{100.000} & 97.619 & 84.171 & \textbf{0.871} \\
    \midrule
    \multirow{6}{*}{VGGT}
      & VGGT                      & 100.000 & 79.630 & 57.937 & 100.000 & 98.413 & 81.993 & 100.000 & 99.471 & 90.503 & 1.766 \\
      & FastVGGT                  & \underline{96.032} & 73.280 & 39.524 & \textbf{100.000} & 97.090 & 74.321 & \textbf{100.000} & 98.677 & 86.305 & 1.823 \\
      & FasterVGGT\_25            & \textbf{100.000} & \textbf{81.481} & \textbf{57.407} & \textbf{100.000} & \textbf{98.413} & \textbf{82.028} & \textbf{100.000} & 99.206 & \underline{90.538} & 1.746 \\
      & FasterVGGT\_75            & \textbf{100.000} & 64.815 & 38.677 & \textbf{100.000} & 93.915 & 68.536 & \textbf{100.000} & 98.942 & 83.016 & 1.372 \\
      & AVGGT(2)                  & \textbf{100.000} & \underline{79.894} & \underline{56.667} & \textbf{100.000} & \underline{98.148} & \underline{81.834} & \textbf{100.000} & \textbf{100.000} & \textbf{90.688} & \underline{1.344} \\
      & AVGGT(4)                  & \textbf{100.000} & 76.190 & 53.492 & \textbf{100.000} & 97.619 & 79.506 & \textbf{100.000} & \underline{99.735} & 89.259 & \textbf{1.207} \\
    \bottomrule
  \end{tabular}
  }
\end{table}

\begin{table}[t]
  \centering
  \caption{
    Camera pose estimation on ScanNet~\cite{dai2017scannet}.
    Best and second best are highlighted within each baseline block, excluding the baseline row.
  }
  \label{tab:sparse_dis_scannetv2}
  \resizebox{\columnwidth}{!}{%
  \begin{tabular}{llcccc}
    \toprule
    Baseline & Method & ATE $\downarrow$ & RPE trans $\downarrow$ & RPE rot $\downarrow$ & Time (s) $\downarrow$ \\
    \midrule
    \multirow{5}{*}{\(\pi^3\)} 
      & \(\pi^3\)                 & 0.030 & 0.013 & 0.346 & 5.718 \\
      & Faster\_\(\pi^3\)\_25     & \textbf{0.030} & \textbf{0.013} & \textbf{0.347} & 5.175 \\
      & Faster\_\(\pi^3\)\_75     & 0.038 & 0.014 & 0.388 & 3.204 \\
      & A\_\(\pi^3\)(2)           & \underline{0.032} & \textbf{0.013} & \underline{0.355} & \underline{2.801} \\
      & A\_\(\pi^3\)(4)           & 0.033 & \textbf{0.013} & 0.365 & \textbf{2.403} \\
    \midrule
    \multirow{6}{*}{VGGT} 
      & VGGT                      & 0.035 & 0.015 & 0.376 & 7.959 \\
      & FastVGGT                  & \underline{0.036} & 0.017 & 0.414 & 4.923 \\
      & FasterVGGT\_25            & \textbf{0.035} & \textbf{0.015} & \textbf{0.379} & 7.238 \\
      & FasterVGGT\_75            & 0.041 & 0.018 & 0.479 & 4.586 \\  
      & AVGGT(2)                  &  \underline{0.036} & \underline{0.016} & \underline{0.384} & \underline{4.376} \\
      & AVGGT(4)                  &  \underline{0.036} & \underline{0.016} & 0.391 & \textbf{3.755} \\
    \bottomrule
  \end{tabular}
  }
\end{table}

\begin{table}[t]
  \centering
  \caption{%
    Point map estimation on NRGBD~\cite{azinovic2022neural}.
    Best and second best are highlighted within each baseline block, excluding the baseline row.
  }
  \label{tab:sparse_point_nrgbd}
  \resizebox{\columnwidth}{!}{%
  \begin{tabular}{llccccccc}
    \toprule
    \multirow{2}{*}{Baseline} & \multirow{2}{*}{Method} &
    \multicolumn{2}{c}{Acc. $\downarrow$} &
    \multicolumn{2}{c}{Comp. $\downarrow$} &
    \multicolumn{2}{c}{NC. $\uparrow$} &
    \multirow{2}{*}{Time (s) $\downarrow$} \\
    \cline{3-4}\cline{5-6}\cline{7-8}
     &  & Mean & Med. & Mean & Med. & Mean & Med. &  \\
    \midrule
    \multirow{5}{*}{\(\pi^3\)}
      & \(\pi^3\)                 & 0.012 & 0.006 & 0.013 & 0.005 & 0.768 & 0.870 & 0.453 \\
      & Faster\_\(\pi^3\)\_25     & \textbf{0.014} & \textbf{0.008} & \textbf{0.014} & \textbf{0.006} & \textbf{0.749} & \textbf{0.865} & 0.493 \\
      & Faster\_\(\pi^3\)\_75     & 0.050 & 0.031 & 0.028 & 0.012 & 0.700 & 0.848 & 0.451 \\
      & A\_\(\pi^3\)(2)           & \underline{0.019} & \underline{0.012} & \underline{0.016} & \underline{0.007} & \underline{0.738} & \textbf{0.865} & \underline{0.411} \\
      & A\_\(\pi^3\)(4)           & 0.028 & 0.019 & 0.020 & 0.009 & 0.732 & 0.864 & \textbf{0.398} \\
    \midrule
    \multirow{6}{*}{VGGT}
      & VGGT                      & 0.013 & 0.007 & 0.015 & 0.006 & 0.784 & 0.877 & 0.619 \\
      & FastVGGT                  & 0.020 & 0.012 & 0.019 & 0.009 & 0.597 & 0.662 & 1.053 \\
      & FasterVGGT\_25            & \textbf{0.014} & \textbf{0.007} & \textbf{0.016} & \textbf{0.006} & \underline{0.776} & \underline{0.875} & 0.637 \\
      & FasterVGGT\_75            & 0.054 & 0.030 & 0.050 & 0.027 & 0.714 & 0.851 & 0.604 \\
      & AVGGT(2)                  & \textbf{0.014} & \textbf{0.007} & \textbf{0.016} & \textbf{0.006} & \textbf{0.781} & \textbf{0.876} & \underline{0.559} \\
      & AVGGT(4)                  & 0.015 & 0.008 & 0.017 & \textbf{0.006} & \underline{0.776} & \underline{0.875} & \textbf{0.555} \\
    \bottomrule
  \end{tabular}
  }
\end{table}

\begin{table}[t]
  \centering
  \caption{
    Ablation results of not subsampling Query tokens for AVGGT on RealEstate10K~\cite{zhou2018stereo}.
    }
  \label{tab:aba_no_QSub}
  \resizebox{\columnwidth}{!}{%
  \begin{tabular}{lccc}
    \toprule
    Method & AUC@5 $\uparrow$ & AUC@15 $\uparrow$ & AUC@30 $\uparrow$ \\
    \midrule
      VGGT         & 63.176 & 81.103 & 88.130 \\
      AVGGT(4)     & 59.583 & 79.188 & 87.045  \\
      AVGGT(2+Q2Near)    & 42.565 & 66.684 & 78.310 \\
      AVGGT(2+Q2GM)          & 26.973 & 53.198 & 68.382 \\
    \bottomrule
  \end{tabular}
  }
\end{table}

\begin{table}[t]
  \centering
  \caption{
    Ablation results of the subsampling strategy for AVGGT on RealEstate10K~\cite{zhou2018stereo}.
    }
  \label{tab:aba_vggt_downsample}
  \resizebox{\columnwidth}{!}{%
  \begin{tabular}{lccc}
    \toprule
    Method & AUC@5 $\uparrow$ & AUC@15 $\uparrow$ & AUC@30 $\uparrow$ \\
    \midrule
      AVGGT(2)     & 61.959 & 80.443 & 87.758 \\
      AVGGT(2\_SIFT)     & 55.438 & 77.415 & 86.034 \\
      AVGGT(2\_Random)     & 57.849 & 78.233 & 86.441 \\
      AVGGT(2\_High)     & 61.220 & 79.982 & 87.448 \\
      AVGGT(2\_Low)     & 59.622       &  79.493      &   87.341       \\
      AVGGT(2\_Mean)   &  56.990      &   77.491     &  85.886        \\
    \bottomrule
  \end{tabular}
  }
\end{table}

\begin{table*}[t]
  \centering
  \caption{
    Ablation results of the diagonal preservation for AVGGT on RealEstate10K~\cite{zhou2018stereo} and 7-Scenes~\cite{shotton2013scene}.
  }
  \label{tab:aba_subsample}
  \begin{tabular}{lcccccc}
    \toprule
    \multirow{2}{*}{Method} &
    \multicolumn{3}{c}{\textbf{RealEstate10K (Sparse)}} &
    \multicolumn{3}{c}{\textbf{7-Scenes (Dense)}} \\
    \cmidrule(lr){2-4} \cmidrule(lr){5-7}
    & AUC@5 $\uparrow$ & AUC@15 $\uparrow$ & AUC@30 $\uparrow$
    & AUC@5 $\uparrow$ & AUC@15 $\uparrow$ & AUC@30 $\uparrow$ \\
    \midrule
    AVGGT(2) &
      61.959 & 80.443 & 87.758 &
      26.061 & 61.951 & 78.113 \\
    AVGGT(2\_WithDiagonal) &
      62.039 & 80.647 & 87.926 &
      24.532 & 60.551 & 77.247 \\
    AVGGT(2\_WithMean) &
      59.427 & 79.030 & 86.930 &
      26.227 & 62.171 & 78.170 \\
    AVGGT(2\_SubsampleOnly) &
      61.908 & 80.539 & 87.838 &
      24.539 & 60.564 & 77.262 \\
    \bottomrule
  \end{tabular}
\end{table*}

\begin{table}[t]
  \centering
  \caption{
    Ablation results on RealEstate10K~\cite{zhou2018stereo} for keeping the first frame tokens in VGGT.
}
  \label{tab:aba_vggt_first}
  \resizebox{\columnwidth}{!}{%
  \begin{tabular}{lccc}
    \toprule
    Method & AUC@5 $\uparrow$ & AUC@15 $\uparrow$ & AUC@30 $\uparrow$ \\
    \midrule
      AVGGT(2)     & 61.959 & 80.443 & 87.758 \\
      AVGGT(2\_FullySubsample) & 60.931 & 79.962 & 87.499 \\
      AVGGT(4)     & 59.583 & 79.188 & 87.045 \\
      AVGGT(4\_FullySubsample) & 55.570 & 76.675 & 85.446 \\
    \bottomrule
  \end{tabular}
  }
\end{table}

\begin{table}[t]
  \centering
  \caption{
    Ablation results on RealEstate10K~\cite{zhou2018stereo} for \(\pi^3\).
  }
  \label{tab:aba_all_pi3}
  \resizebox{\columnwidth}{!}{%
  \begin{tabular}{lccc}
    \toprule
    Method & AUC@5 $\uparrow$ & AUC@15 $\uparrow$ & AUC@30 $\uparrow$ \\
    \midrule
      \(\pi^3\)         & 67.186 & 83.288 & 89.500 \\
      \(\pi^3\)(G2F)      & 66.790 & 83.006 & 89.313 \\
      \(\pi^3\)(G2M)       & 63.166 & 80.593 & 87.766 \\
      A\_\(\pi^3\)(2)     & 64.832 & 82.306 & 89.008 \\
      A\_\(\pi^3\)(4)     & 58.703 & 79.350 & 87.434 \\
      A\_\(\pi^3\)(6)     & 51.780 & 75.037 & 84.718 \\
      A\_\(\pi^3\)(9)     & 40.437 & 65.927 & 78.060 \\
    \bottomrule
  \end{tabular}
  }
\end{table}

\begin{figure}
\centering
\resizebox{\columnwidth}{!}{\includegraphics{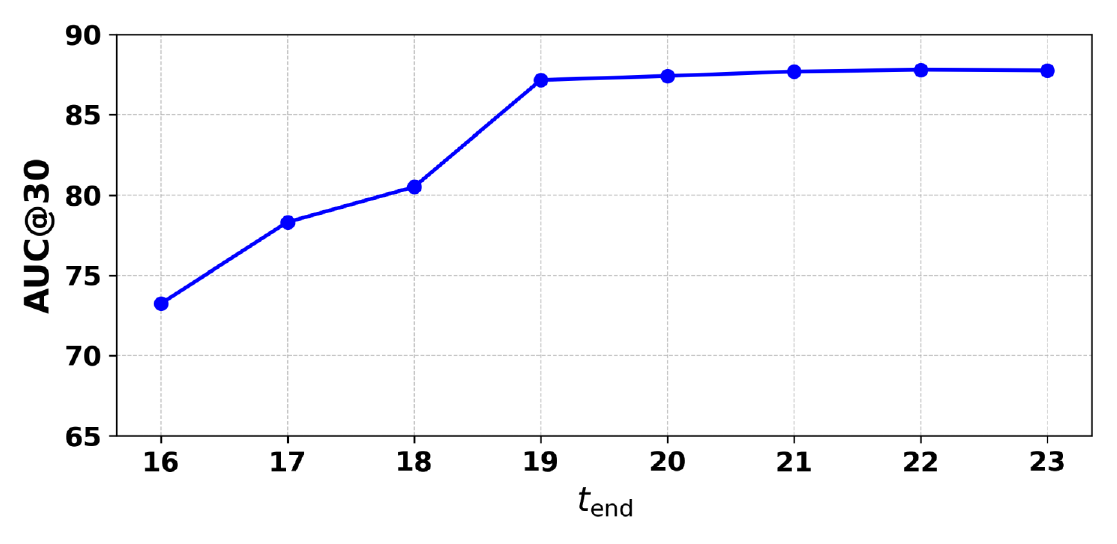}}
\caption{Ablation results on RealEstate10K for different VGGT \(t_{end}\) choices.}
\label{fig:tend_ablation}
\end{figure}

\begin{figure}
\centering
\resizebox{\columnwidth}{!}{\includegraphics{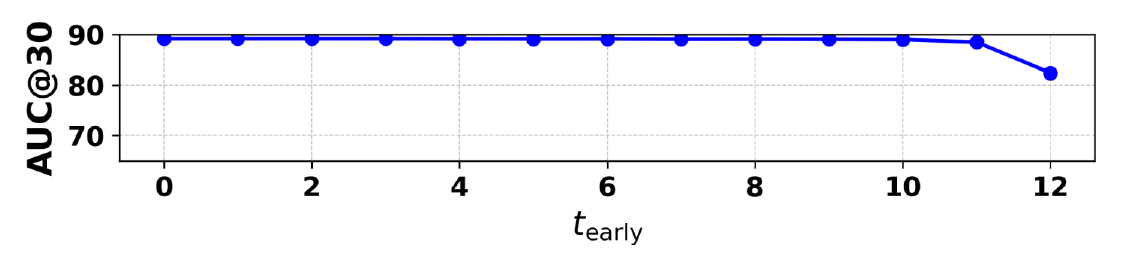}}
\caption{Ablation results on RealEstate10K for different \(\pi^3\) \(t_{early}\) choices.}
\label{fig:pi3_tearly}
\end{figure}

\begin{figure}
\centering
\resizebox{\columnwidth}{!}{\includegraphics{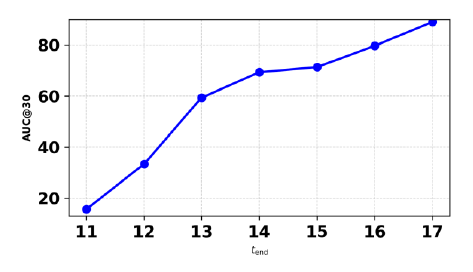}}
\caption{Ablation results on RealEstate10K for different \(\pi^3\) \(t_{end}\) choices.}
\label{fig:pi3_tend}
\end{figure}

\section{More Analysis}

\subsection{Analyzing $\pi^3$ Global Attention}
Following the analysis conducted for VGGT~\cite{wang2025vggt} in the main paper, we first briefly introduce the $\pi^3$~\cite{wang2025pi} architecture. Given multiple input images, $\pi^3$ uses a frozen DINOv2~\cite{oquab2023dinov2} encoder. For each frame, five learnable register tokens~\cite{darcet2023vision} are appended to the patch tokens. All frame tokens are then passed through an aggregator composed of 36 transformer blocks that alternate between frame attention and global attention. After aggregation, all register tokens are removed, and the remaining tokens are fed into the camera and point heads to predict camera poses and point clouds.
Unlike VGGT, $\pi^3$ discards camera tokens and adopts a fully permutation-equivariant architecture with respect to the input frames.

\begin{figure*}[p]
\centering
\resizebox{0.95\textwidth}{!}{\includegraphics{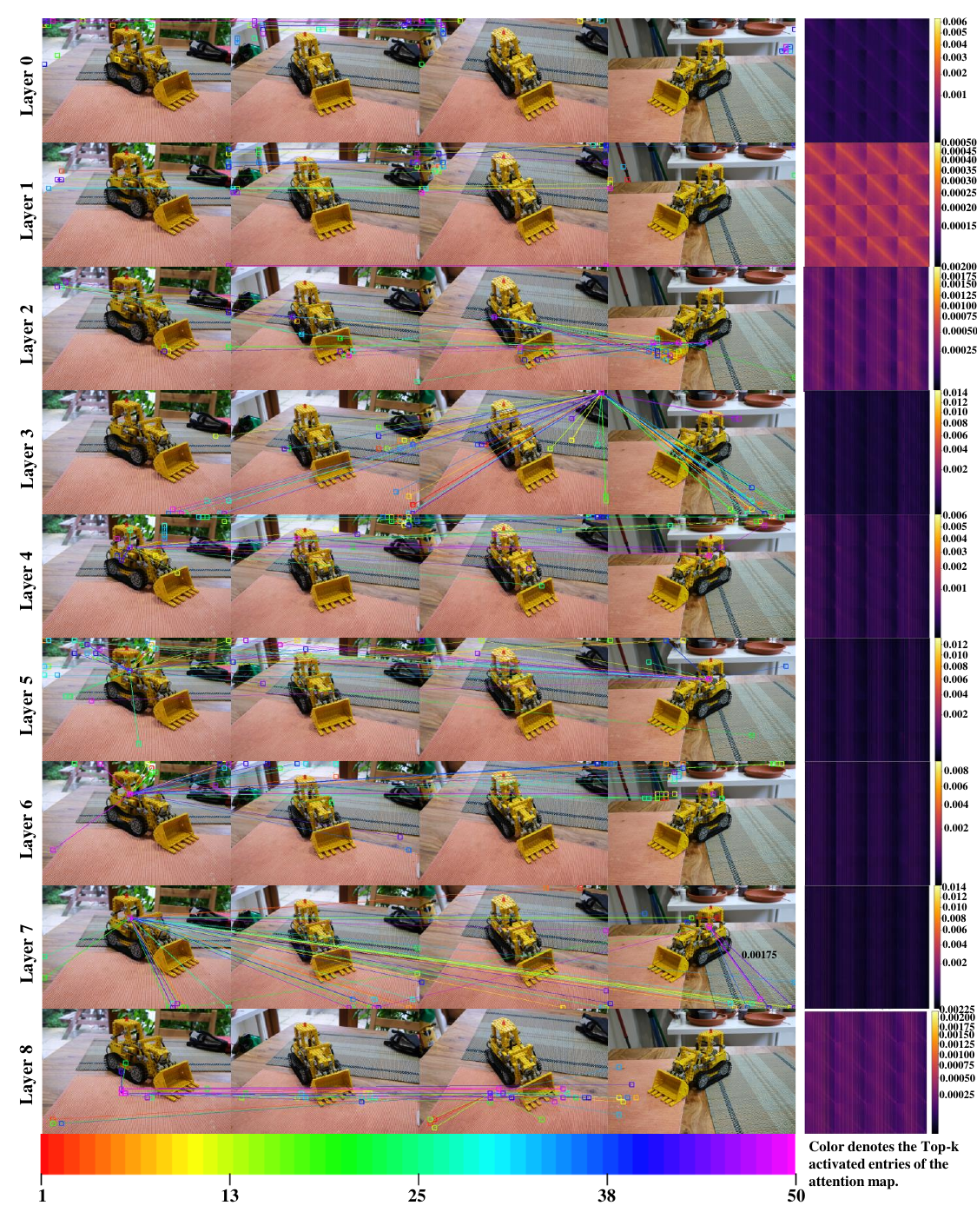}}
\caption{Visualization of global attention for layers 0-8 in \(\pi^3\).}
\label{fig:pi3_1}
\end{figure*}

\begin{figure*}[p]
\centering
\resizebox{0.95\textwidth}{!}{\includegraphics{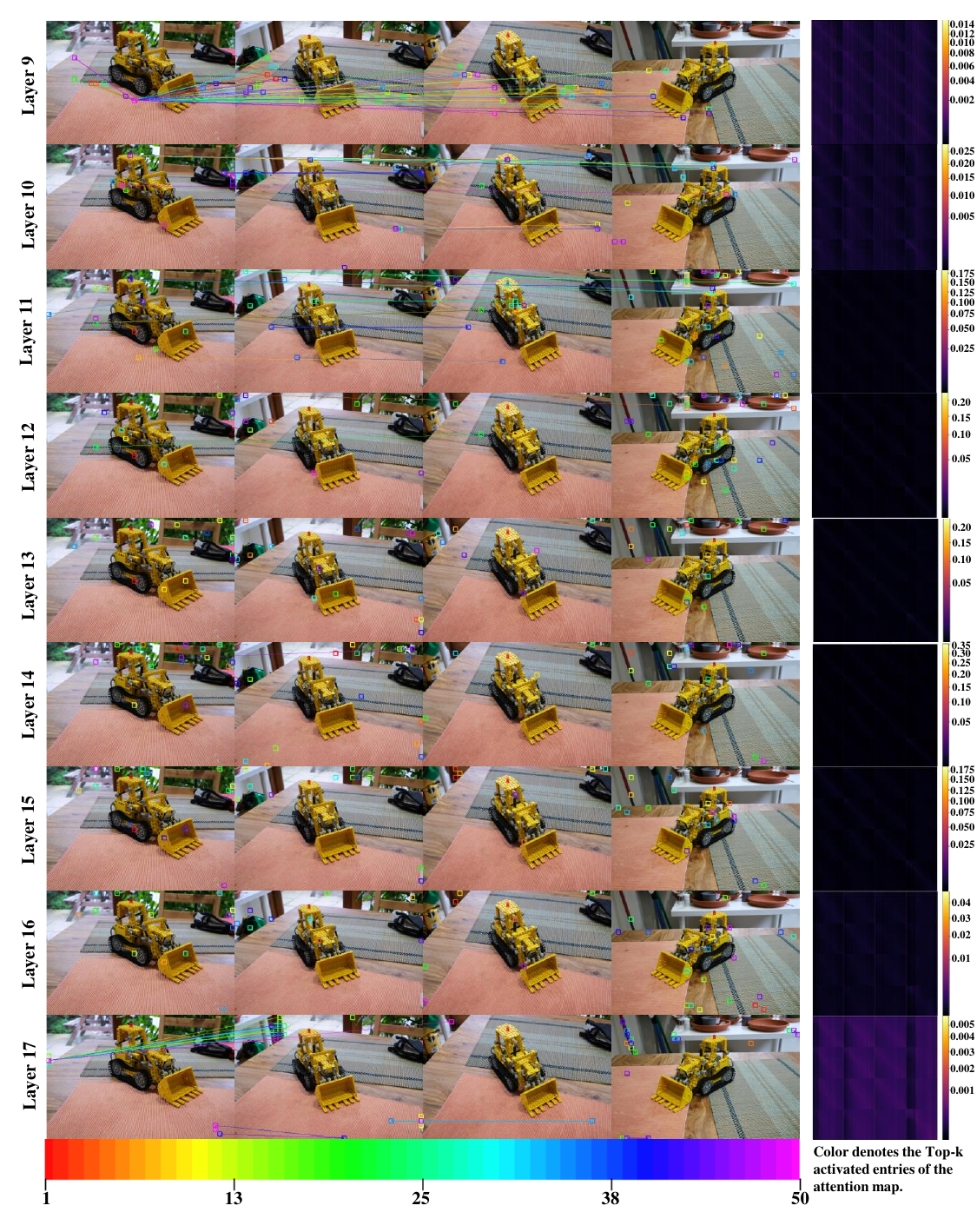}}
\caption{Visualization of global attention for layers 9-17 in \(\pi^3\).}
\label{fig:pi3_2}
\end{figure*}

We visualize all global attention layers (indices 0-17) in Fig.~\ref{fig:pi3_1} and Fig.~\ref{fig:pi3_2}, and analyze four representative layers (indices 1, 3, 11, and 17). Overall, the observations are highly consistent with those for VGGT. In the early global attention layers (indices 0-9), the maximum attention values are significantly smaller than those in the middle layers, indicating a more uniform distribution. The top activated entries reveal two characteristic patterns: in the very first layers (\textit{e.g.}, layers 0-1), attention is dominated by positional embeddings, whereas in layers 2-9, attention frequently focuses on a small and inconsistent subset of key tokens. These behaviors suggest that the early global attention layers contribute little to establishing cross-view correspondences.

In the middle global attention layers (indices 10-16), the attention becomes far more selective, with noticeably larger peak values. The highest responses predominantly correspond either to self-attention on the same patch or to cross-view patches at the same spatial location, indicating that these layers are responsible for building cross-view correlations. Finally, the last global attention layer (index 17) again tends toward a more uniform distribution. However, unlike in VGGT, the top-activated token pairs in this layer still exhibit clear 3D-related structure. Therefore, in $\pi^3$, we consider that the last global attention layer still contributes to building cross-view correspondences.

\subsection{Rotation Test for Early Global Attention Layers}
For the early global attention layers in VGGT/$\pi^3$, the attention matrices exhibit nearly uniform distributions, and the top activated entries frequently attend to a small subset of key tokens. To investigate whether this subset encodes any meaningful 3D or view-consistent information, we conduct the following test. Instead of directly feeding the original input images, we rotate all input views by $180^\circ$ and then pass them through VGGT. This operation is equivalent to rotating the camera while keeping the underlying 3D scene unchanged.

As illustrated in Fig.~\ref{fig:rot_ablation}, we first collect the top 1000 attention entries (over all query–key pairs) in both the original and rotated runs, respectively. We then identify the entries that share the same query patch token in the two runs and, following the attention ranking in the original run, select the top 50 such entries for analysis. For visualization clarity, we rotate the outputs back to the original orientation. We observe that, for early global attention layers (\textit{e.g.}, Layer~4), the highly activated key tokens change almost entirely after rotation. In contrast, for middle global attention layers (\textit{e.g.}, Layer~11), the corresponding entries in the rotated case still largely point to the 
same spatial locations across both runs. 
This indicates that the highly activated tokens in the early global attention layers do not correspond to stable 3D structures or view-consistent relationships. Therefore, this rotation test further supports our conclusion that the early global attention layers do not contribute meaningfully to building cross-view correspondences.

\section{More Experiments}
\subsection{Additional Dataset Results}
Here, we provide additional results on more datasets. 
For camera pose estimation, we further evaluate on ScanNet~\cite{dai2017scannet} and Sintel~\cite{bozic2021transformerfusion} and ETH3D~\cite{schops2017multi}. 
For point-map estimation, we additionally report results on NRGBD~\cite{azinovic2022neural}. 
The results are summarized in Tables~\ref{tab:sparse_dis_sintel},~\ref{tab:ETH3D},~\ref{tab:sparse_dis_scannetv2} and~\ref{tab:sparse_point_nrgbd}.
We observe that on several datasets, FasterVGGT achieves better metrics than our method; however, this is mainly because its sparse ratio is significantly smaller than our minimum subsampling factor (a factor of 2 corresponds to keeping 50\% of patch tokens), resulting in substantially higher computational cost. 
Under comparable runtime budgets, our method attains superior accuracy. 
Overall, our approach consistently achieves the best trade-off between accuracy and efficiency across all evaluated datasets.

\subsection{More Ablation Studies on VGGT}
\paragraph{Effect of Not Subsampling Query Tokens.}
We evaluated two training-free Query-subsampling variants on top of AVGGT(2) (with K/V already subsampled using $\sigma{=}2$):
\textbf{AVGGT(2+Q2GM)} uses a global-mean Query token to approximate the missing attention weights induced by Query subsampling;
\textbf{AVGGT(2+Q2Near)} similarly replaces each skipped Query with its nearest retained Query, without additional training.
In terms of computational complexity, both AVGGT(2+Q2GM) and AVGGT(2+Q2Near) are comparable to AVGGT(4).
We evaluate them on RealEstate10K. As shown in Table~\ref{tab:aba_no_QSub}, both variants perform substantially worse than AVGGT(4).
We believe this performance drop arises because, while global attention is functionally used to build cross-view correspondences, its effect is ultimately realized through attention computation that updates token values. Consequently, subsampling Query tokens inevitably reduces differences between tokens, which is clearly harmful for 3D tasks that require dense features.
This interpretation is consistent with the results: AVGGT(2+Q2GM) performs worse than AVGGT(2+Q2Near), since using a global-mean Query makes different tokens even more similar.

\paragraph{Effect of the Grid-Based Subsampling Strategy.}
Our subsampling strategy is grid-based, but we also explored several alternative token-selection methods. 
Since our analysis shows that global attention builds cross-view correspondences by aligning tokens at the same spatial positions, we first draw inspiration from traditional SfM pipelines~\cite{cui2017hsfm,hartley2003multiple,pan2024global} such as COLMAP~\cite{schonberger2016structure,schonberger2016pixelwise}, where SIFT~\cite{lowe2004distinctive} keypoints are used for feature matching. 
Although patch tokens and SIFT keypoints are intrinsically different, we test whether SIFT-based cues can help select more informative tokens. 
Specifically, we detect SIFT keypoints on all input images, accumulate their scores over the patch grid, and select the top-scoring patch tokens. 
We denote this variant as \text{AVGGT(2\_SIFT)}, meaning a subsampling factor of 2 with SIFT-based token selection. 
As shown in Tab.~\ref{tab:aba_vggt_downsample}, this variant performs worse than our grid-based strategy, suggesting that the tokens used by global attention for alignment do not align well with traditional handcrafted keypoint pipelines.
Within the grid-based setting, we further evaluate whether selecting a fixed spatial index is optimal. 
We consider four variants: randomly selecting a token within each grid cell (\text{AVGGT(2\_Random)}), selecting the token with the highest feature magnitude (\text{AVGGT(2\_High)}), selecting the lowest-magnitude token (\text{AVGGT(2\_Low)}), and using the mean value within each cell (\text{AVGGT(2\_Mean)}). 
In \text{AVGGT(2\_Mean)}, we exclude both the diagonal and the mean components from our enhanced strategy.
As shown in Tab.~\ref{tab:aba_vggt_downsample}, the fixed grid-based selection remains the best-performing choice, whereas \text{AVGGT(2\_Mean)} performs the worst. 
We attribute this to the fact that global attention constructs cross-view correspondences through spatially consistent tokens, while mean aggregation destroys these spatial anchors that are crucial for cross-view alignment.

\paragraph{Effect of Diagonal Preservation.}
We also study our enhanced subsampling strategy, which incorporates diagonal preservation and a mean component. We denote the variant without either enhancement (with subsampling factor 2) as AVGGT(2\_SubsampleOnly), the variant with only the mean component as AVGGT(2\_WithMean), and the variant with only diagonal preservation as AVGGT(2\_WithDiagonal). As shown in Tab.~\ref{tab:aba_subsample}, diagonal preservation yields a slight improvement in the sparse setting but slightly degrades performance under dense inputs, while the mean component exhibits the opposite trend. Despite these differences, combining both enhancements achieves the best overall performance, which is why we adopt this configuration as our default choice.

\paragraph{Effect of the Last Global Attention Layers.}
In the previous ablation studies, we observed that directly replacing the last global attention layers (indices 20–23) with frame attention results in only a slight performance drop. Here, we provide a more detailed analysis by introducing a parameter \(t_{end}\), which specifies that all global attention layers with indices \(> t_{end}\) are converted to frame attention. As shown in Fig.~\ref{fig:tend_ablation}, the last global attention layers contribute very little to the final performance, and this effect becomes even more pronounced as \(t_{end}\) increases. We interpret this behavior as evidence that, at deeper layers, the global feature maps are already well aligned across views, so additional global attention contributes less to cross-view consistency, leading to minimal performance difference.

\paragraph{Effect of Not Subsampling the First Frame.}
Our subsampling strategy differs slightly between VGGT and $\pi^3$. Since VGGT treats the first frame as the reference view, this frame generally plays a more important role than the others. Therefore, for VGGT we choose not to subsample the patch tokens of the first frame. We denote by AVGGT(2\_FullySubsample) the variant that subsamples all frames with a factor of 2. As shown in Tab.~\ref{tab:aba_vggt_first}, keeping the first frame uncompressed yields a small but consistent performance improvement.

\subsection{More Ablation Studies on $\pi^3$}
Similar to our ablation results on VGGT, as shown in Fig.~\ref{fig:pi3_tearly}, the early global attention layers in $\pi^3$ do not contribute to building cross-view correspondences. We observe that \(t_{early}=10\) provides the best trade-off between accuracy and speed. From Tab.~\ref{tab:aba_all_pi3}, both \(\pi^3\)(G2F) and \(\pi^3\)(G2M) modify global attention layers with indices 0-9. The fact that \(\pi^3\)(G2F) achieves nearly the same performance as the original model indicates that no cross-view correspondences are formed in these layers, while the small performance gap of \(\pi^3\)(G2M) further suggests that these layers contain little meaningful selective attention. Regarding the subsampling factor, $\pi^3$ follows the same pattern as VGGT: it is more sensitive under sparse inputs and more robust when the input views are dense.
A difference arises in the behavior of the last global attention layers. As shown in Fig.~\ref{fig:pi3_tend}, in $\pi^3$ even Layer~17 still has a noticeable impact on the final performance. We attribute this to architectural differences: $\pi^3$ contains only 36 alternating transformer blocks, whereas VGGT contains 48. From this perspective, $\pi^3$ likely has less redundancy in its deeper layers, making its final global attention layer more influential than in VGGT.

\section{More Visualization Results of Global Attention for VGGT}
\begin{figure*}[p]
\centering
\resizebox{\textwidth}{!}{\includegraphics{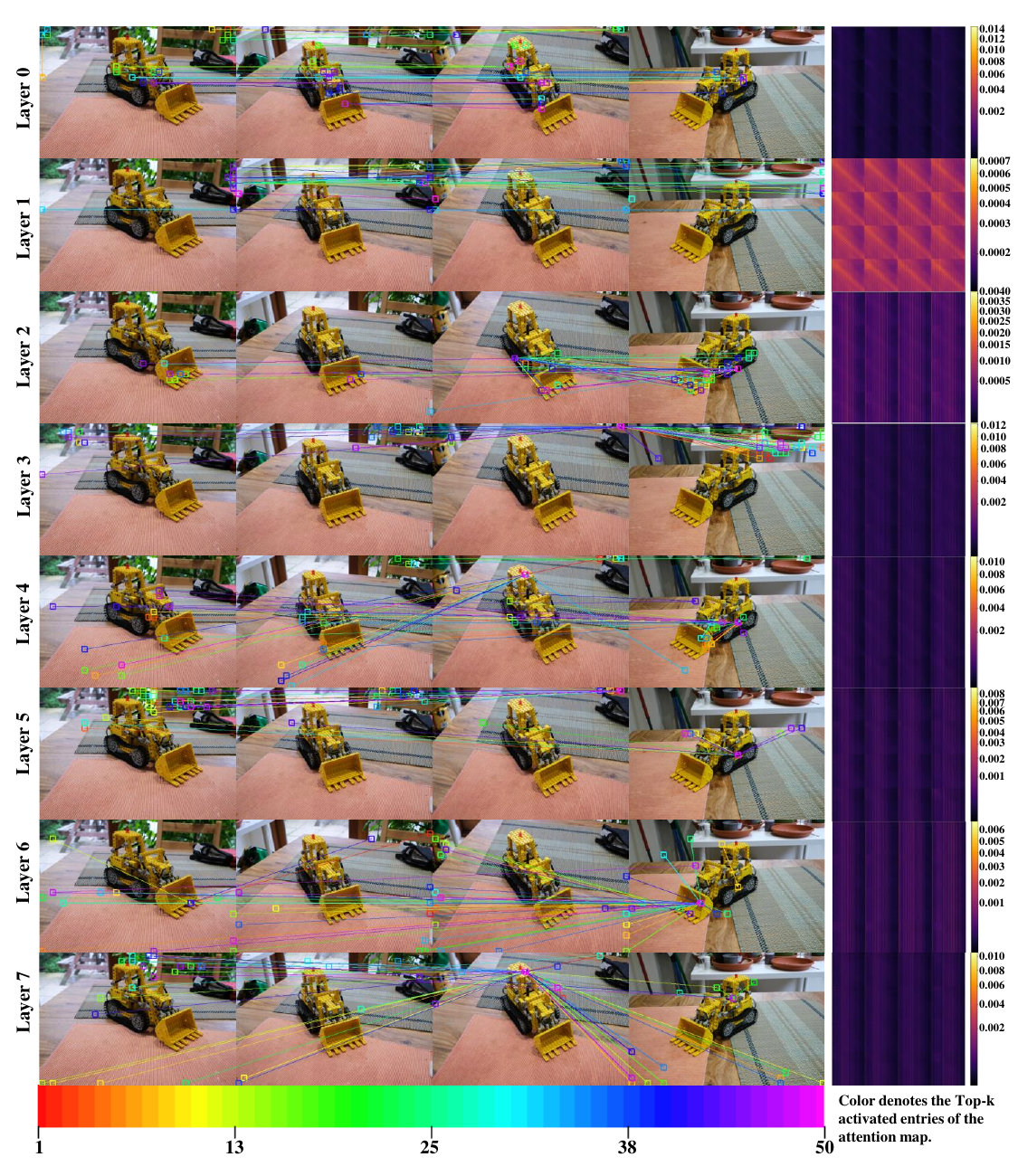}}
\caption{Visualization of global attention for layers 0-7 in VGGT.}
\label{fig:vggt1}
\end{figure*}

\begin{figure*}[p]
\centering
\resizebox{\textwidth}{!}{\includegraphics{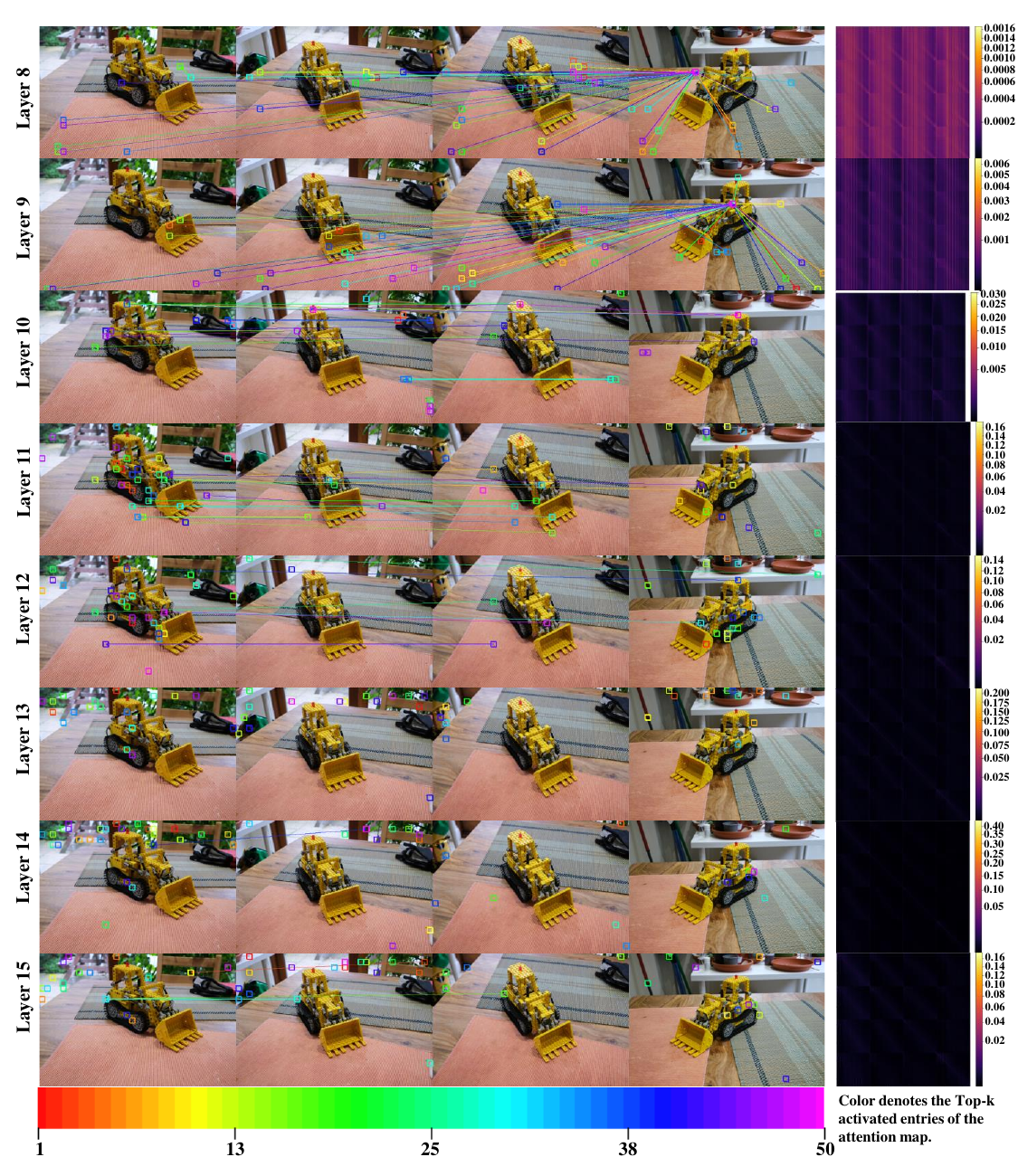}}
\caption{Visualization of global attention for layers 8-15 in VGGT.}
\label{fig:vggt2}
\end{figure*}

\begin{figure*}[p]
\centering
\resizebox{\textwidth}{!}{\includegraphics{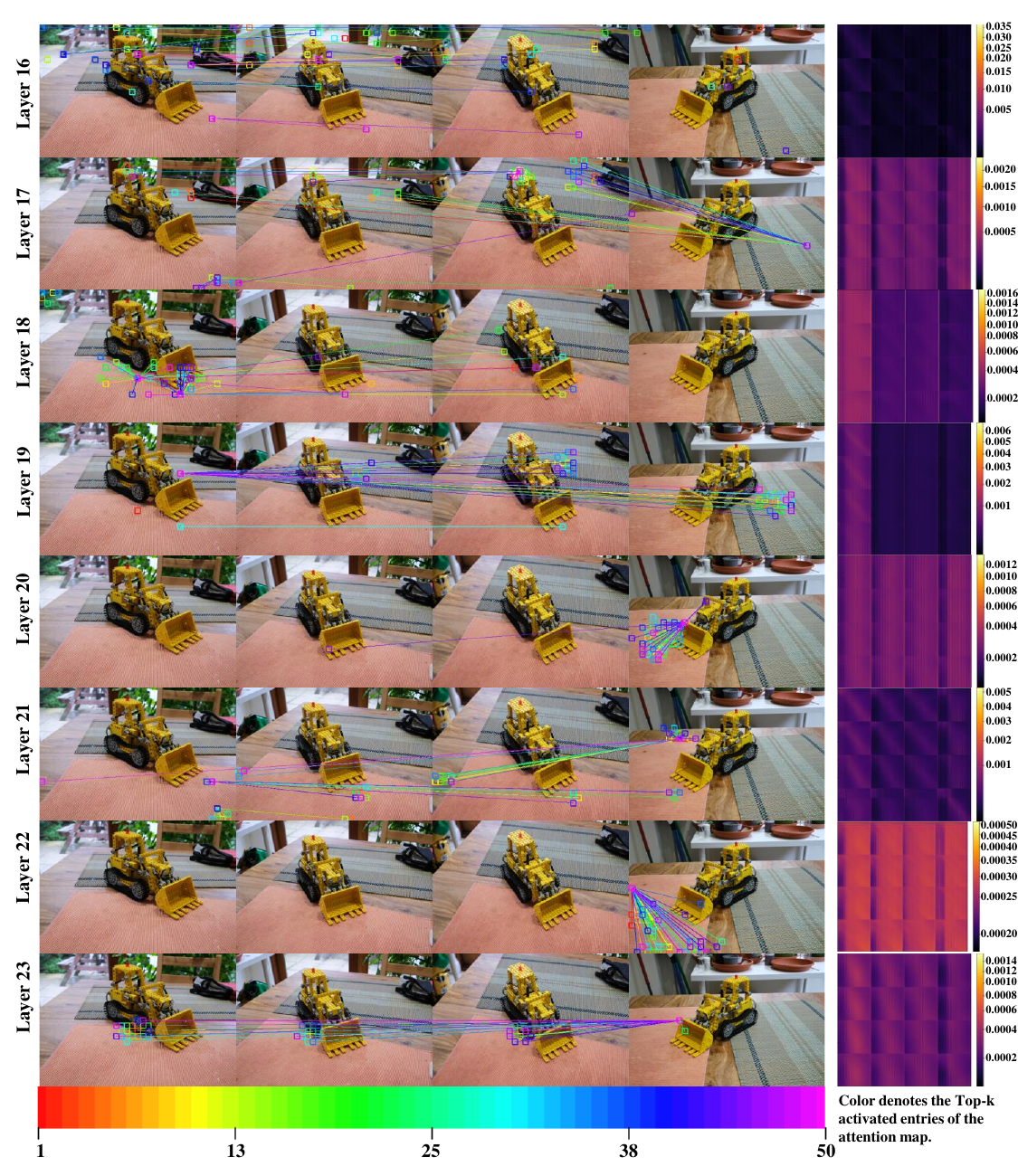}}
\caption{Visualization of global attention for layers 16-23 in VGGT.}
\label{fig:vggt3}
\end{figure*}

\begin{figure*}[p]
\centering
\resizebox{\textwidth}{!}{\includegraphics{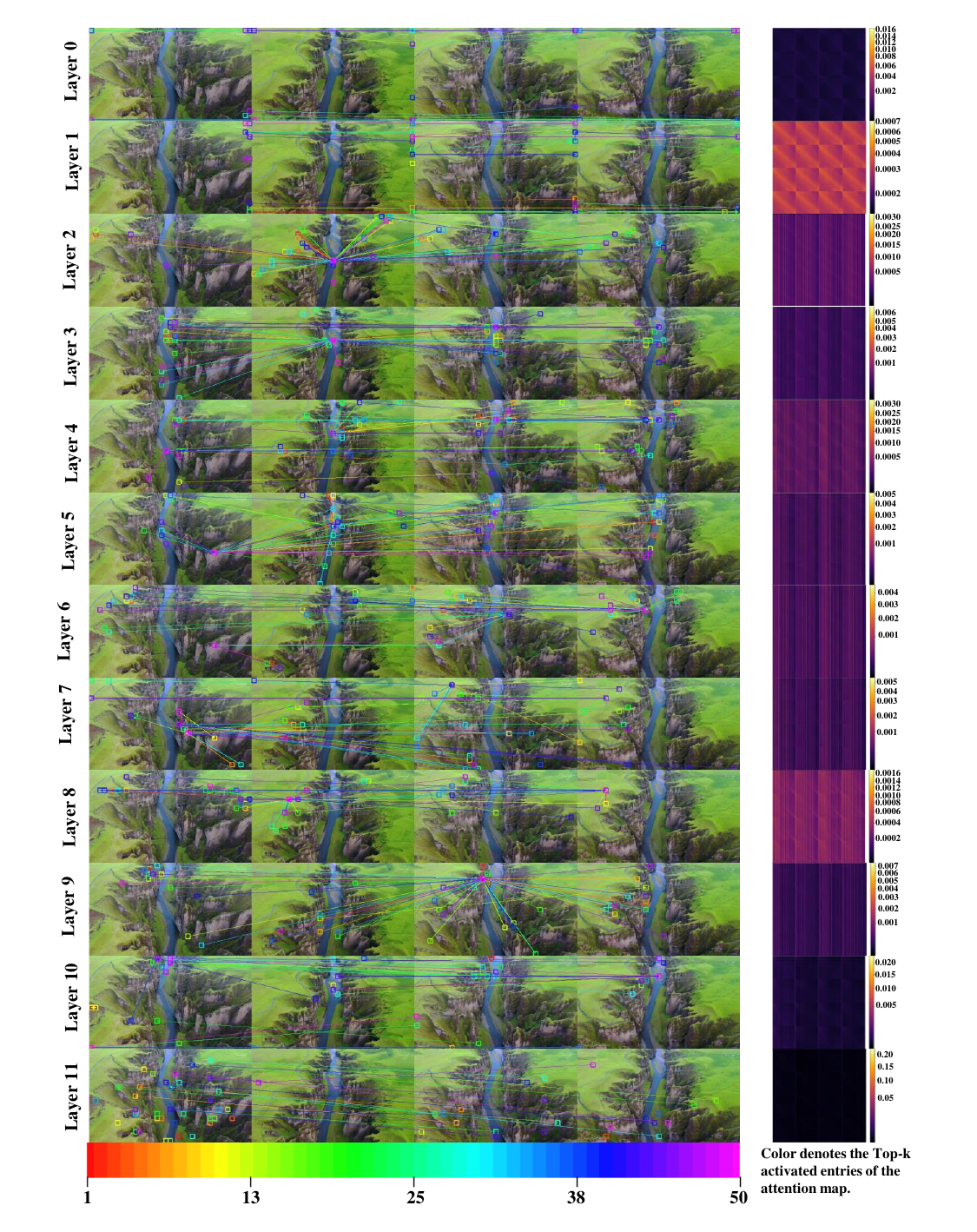}}
\caption{Visualization of global attention for layers 0-11 in VGGT.}
\label{fig:vggt1_1}
\end{figure*}

\begin{figure*}[p]
\centering
\resizebox{\textwidth}{!}{\includegraphics{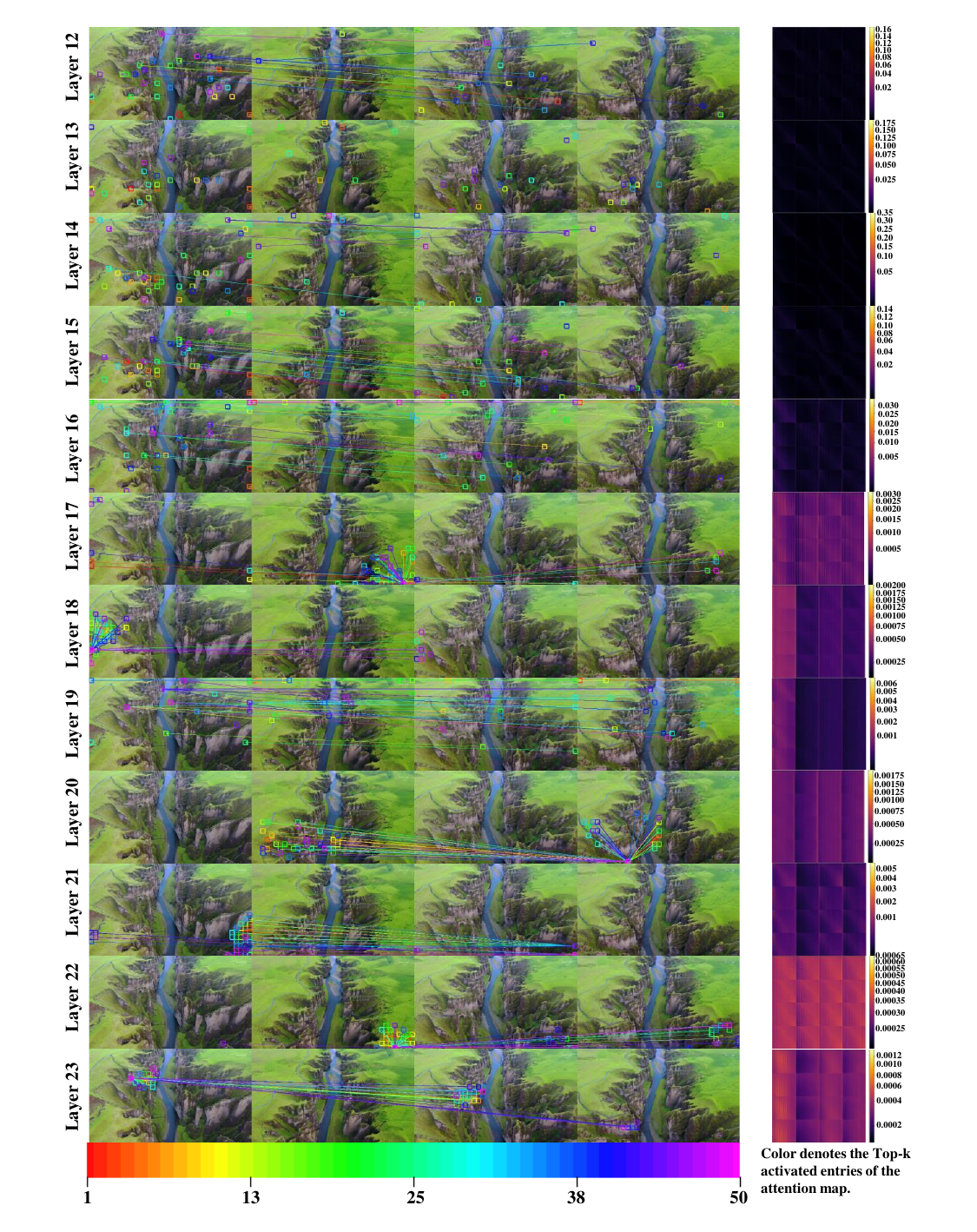}}
\caption{Visualization of global attention for layers 12-23 in VGGT.}
\label{fig:vggt1_2}
\end{figure*}

\begin{figure*}[p]
\centering
\resizebox{\textwidth}{!}{\includegraphics{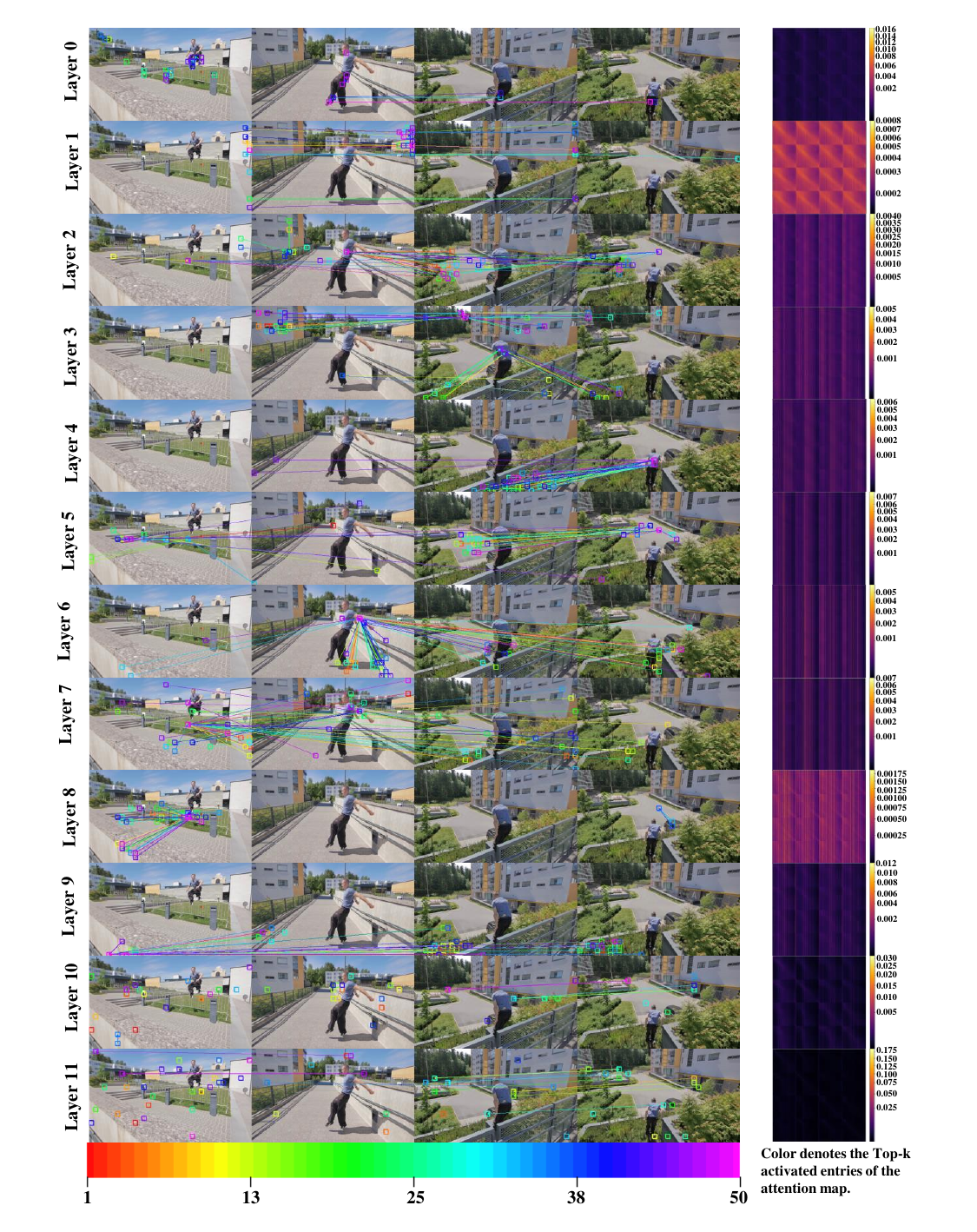}}
\caption{Visualization of global attention for layers 0-11 in VGGT.}
\label{fig:vggt2_1}
\end{figure*}

\begin{figure*}[p]
\centering
\resizebox{\textwidth}{!}{\includegraphics{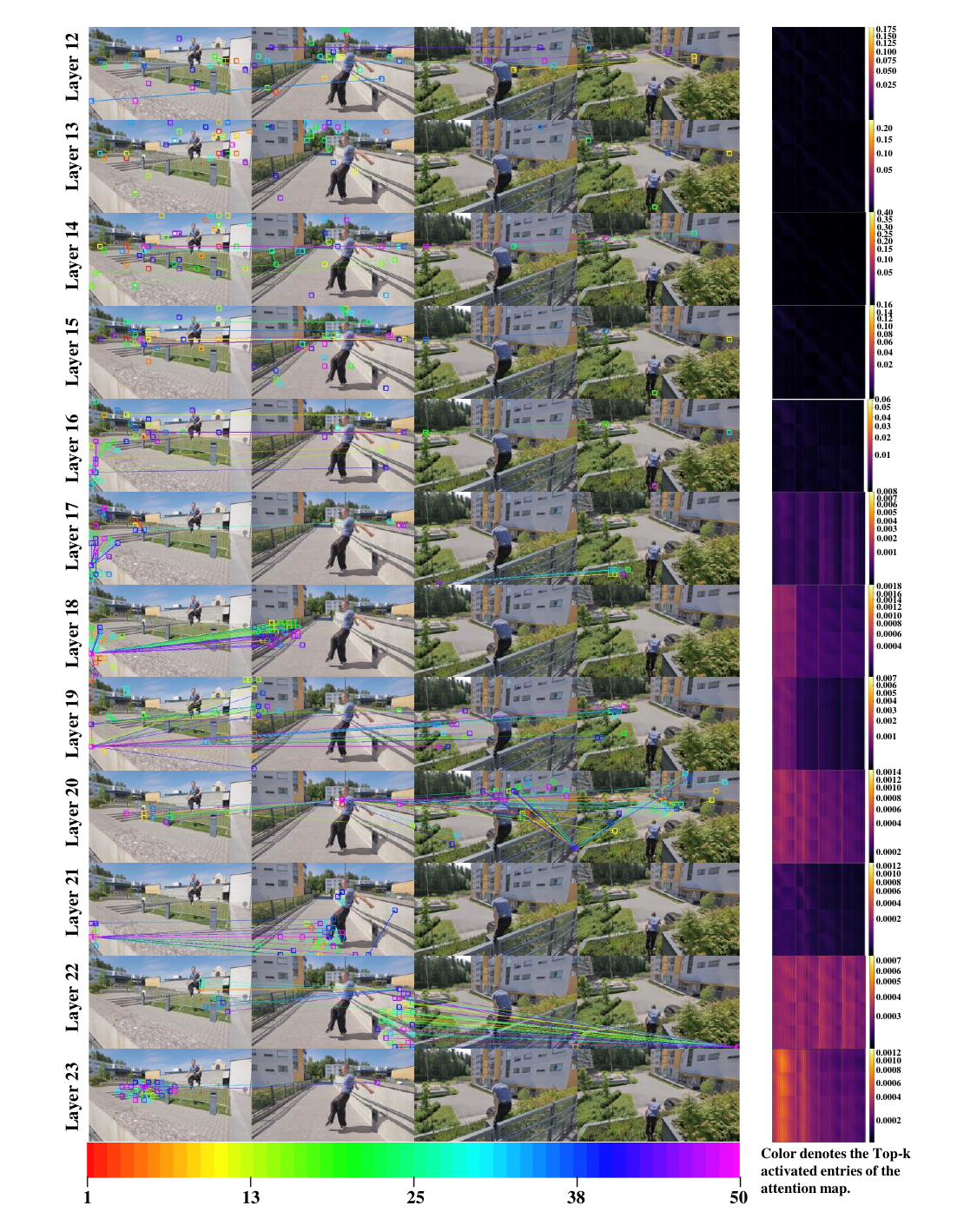}}
\caption{Visualization of global attention for layers 12-23 in VGGT.}
\label{fig:vggt2_2}
\end{figure*}

In addition to Figs.~\ref{fig:vggt1}–\ref{fig:vggt3}, we provide further visualizations of the global attention layers in VGGT. Please refer to Figs.~\ref{fig:vggt1_1}, \ref{fig:vggt1_2}, \ref{fig:vggt2_1}, and \ref{fig:vggt2_2}.

